%% file: main.tex
\RequirePackage{snapshot}
\documentclass[10pt,twocolumn,letterpaper]{article}

\usepackage{cvpr}              %

\input{preamble}

\definecolor{cvprblue}{rgb}{0.21,0.49,0.74}
\usepackage[pagebackref,breaklinks,colorlinks,citecolor=cvprblue]{hyperref}

\def\paperID{370} %
\def\confName{3DV\xspace}
\def\confYear{2024\xspace}

\newcommand\blfootnote[1]{%
  \begingroup
  \renewcommand\thefootnote{}\footnote{#1}%
  \addtocounter{footnote}{-1}%
  \endgroup
}

\title{Objects With Lighting: A Real-World Dataset for Evaluating \\Reconstruction and Rendering for Object Relighting}

\author{%
Benjamin Ummenhofer\textsuperscript{1,*}\\[-5.0mm]
\and
Sanskar Agrawal\textsuperscript{2}
\and
Rene Sepulveda\textsuperscript{1}
\and
Yixing Lao\textsuperscript{3}
\and
Kai Zhang\textsuperscript{4}
\and
Tianhang Cheng\textsuperscript{5}
\and
Stephan Richter\textsuperscript{1}
\and
Shenlong Wang\textsuperscript{5}
\and
German Ros\textsuperscript{6,\dag}\\[-4.8mm]
\and
\textsuperscript{1}Intel Labs,
\textsuperscript{2}Preimage,
\textsuperscript{3}The University of Hong Kong,
\textsuperscript{4}Adobe,\\
\textsuperscript{5}University of Illinois Urbana-Champaign,
\textsuperscript{6}NVIDIA%
}

\begin{document}
\maketitle
\blfootnote{\textsuperscript{*}Corresponding author email  benjamin.ummenhofer@intel.com}
\blfootnote{\textsuperscript{\dag}Work done while working at Intel Labs.}

\begin{abstract}
\input{abstract}
\end{abstract}
\vspace{-5mm}

\input{intro_v2}

\input{related_work_v2}

\input{dataset}

\input{method}

\input{experiments}

\input{limitations}

\input{conclusion}

{
    \small
    \bibliographystyle{ieeenat_fullname}
    \bibliography{main}
}
\clearpage
\setcounter{page}{1}
\maketitlesupplementary
\appendix

\input{appendix_dataset}

\input{appendix_experiments}

\input{appendix_datasheet}

\end{document}

%% file: preamble.tex
\usepackage[dvipsnames]{xcolor}

\usepackage[utf8]{inputenc} %
\usepackage[T1]{fontenc}    %
\usepackage{url}            %
\usepackage{booktabs}       %
\usepackage{amsfonts}       %
\usepackage{nicefrac}       %
\usepackage{microtype}      %
\usepackage{xcolor}         %
\usepackage{caption}
\usepackage{subcaption}

\usepackage{amsmath,amssymb} %
\usepackage{adjustbox}[Export]
\usepackage{dirtree}
\usepackage{array}
\newcolumntype{H}{>{\setbox0=\hbox\bgroup}c<{\egroup}@{}}
\usepackage{stackengine}
\usepackage{wrapfig}

\input{vcl-shortcuts.tex}
\input{shortcuts.tex}

\definecolor{questionblue}{RGB}{80,80,205}
\newcommand{\Q}[1]{\vspace{1ex}\noindent{\color{questionblue}\bfseries#1}\\[1ex]\noindent}

%% file: vcl-shortcuts.tex
\usepackage{epsfig}
\usepackage{graphicx}
\usepackage{float}
\usepackage{wrapfig}
\usepackage{algorithm,algorithmicx,algpseudocode}
\usepackage{bm,xspace}
\usepackage{multirow}
\usepackage{balance}
\usepackage{url}
\usepackage{booktabs}
\usepackage{etoolbox,siunitx}
\usepackage{multirow}
\usepackage[normalem]{ulem}
\usepackage{calc}
\usepackage{pifont,hologo}
\usepackage{nicefrac}

\usepackage[super]{nth}
\usepackage{nicefrac}
\robustify\bfseries

\def\nn{\mathbf{n}}

\def\tt{\mathbf{t}}

\def\vv{\mathbf{v}}

\def\xx{\mathbf{x}}

\def\KK{\mathbf{K}}

\def\RR{\mathbf{R}}

\def\XX{\mathbf{X}}

\def\lL{\mathcal{L}}

\DeclareMathSymbol{@}{\mathord}{letters}{"3B}

\newcommand\mypara[1]{\noindent\textbf{#1}}

\def\latex/{\LaTeX}
\def\bibtex/{\hologo{BibTeX}}

\def\bn{\mathbf{n}} %

\def\bomega{\bm{\omega}} %

%% file: shortcuts.tex
\newcommand{\fig}[1]{\mbox{Fig.\ \ref{#1}}}
\newcommand{\tab}[1]{\mbox{Table~\ref{#1}}}
\newcommand{\sectn}[1]{\mbox{Section~\ref{#1}}}

%% file: abstract.tex
Reconstructing an object from photos and placing it virtually in a new environment goes beyond the standard novel view synthesis task as the appearance of the object has to not only adapt to the novel viewpoint but also to the new lighting conditions and yet evaluations of inverse rendering methods rely on novel view synthesis data or simplistic synthetic datasets for quantitative analysis.
This work presents a real-world dataset for measuring the reconstruction and rendering of objects for relighting.
To this end, we capture the environment lighting and ground truth images of the same objects in multiple environments allowing to reconstruct the objects from images taken in one environment and quantify the quality of the rendered views for the unseen lighting environments.
Further, we introduce a simple baseline composed of off-the-shelf methods and test several state-of-the-art methods on the relighting task and show that novel view synthesis is not a reliable proxy to measure performance.
Code and dataset are available at \adjustbox{width=\columnwidth}{\url{https://github.com/isl-org/objects-with-lighting}}.

%% file: intro_v2.tex
\section{Introduction}
\label{sec:intro}

\textit{Inverse rendering}, the task of recovering an object's shape and material appearance from a collection of images, has been a longstanding goal in the fields of computer vision and graphics~\cite{barrow1978recovering, yu1999inverse}. Its potential applications span across diverse fields, such as photorealistic rendering~\cite{zhangNeRFactorNeuralFactorization2021, bossNeuralPILNeuralPreIntegrated2021}, video editing~\cite{guo2019relightables}, simulation~\cite{straub2019replica, chen2021geosim} and mixed reality~\cite{karsch2011rendering, zhangPhySGInverseRendering2021}. However, inverse rendering poses significant challenges due to its ill-posed nature. The color of a pixel depends on a wide range of factors — often more numerous than the pixels observed — including but not limited to surface material, illumination, viewing angle, and surface geometry.

Numerous methods have been proposed to tackle this problem, ranging from data-driven learning-based methods~\cite{matusik2003data, nishino2009directional, kautz2002fast, lombardi2015reflectance, zhangPhySGInverseRendering2021} to model-driven, optimization-based approaches~\cite{barron2015shape, zhangNeRFactorNeuralFactorization2021, bossNeuralPILNeuralPreIntegrated2021, huDeepBRDFDeepRepresentation2020, sztrajmanNeuralBRDFRepresentation2021, chenInvertibleNeuralBRDF2020}. These have shown promising results in creating digital replicas of objects and rendering them from novel viewpoints and under different lighting conditions. Although qualitative comparisons among these methods are commonly made, the establishment of a comprehensive, unbiased, quantitative benchmark for evaluating inverse graphics approaches remains an elusive goal. This is largely because measuring "ground truth" surface reflective materials and illumination in the wild is notoriously difficult. Existing methods often resort to synthetic data~\cite{zhang2022invrender, liu2023nero} or controlled lab settings~\cite{MurmannGAD19, jensen2014large, toschi2023rene}, which inevitably restrict the applicability of these findings to real-world inverse graphics due to the domain shift. 
Merely evaluating novel view synthesis is inadequate since one can effectively render an object from a novel viewpoint without comprehending its materials and illumination. Instead, it becomes crucial to determine how effectively a relightable digital replica can generate realistic, high-fidelity images under diverse and novel lighting conditions. Despite this need, to the best of our knowledge, a well-calibrated, real-world, multi-illumination dataset enabling quantitative evaluation of object relighting remains absent.

Motivated by these challenges, we present a novel, naturalistic, real-world dataset and benchmark for object reconstruction, rendering, and relighting. This dataset includes ground-truth environmental lighting and posed image observations for each object, captured across various natural illumination environments and viewpoints. All images have been geometrically and photometrically calibrated, facilitating the reconstruction of an object's shape, material, and illumination from a single lighting environment, the object's relighting under unseen lighting conditions, and the measurement of realism and fidelity against reference image observations. The dataset is characterized by a broad diversity of objects and illumination environments, spanning from non-reflective to reflective surfaces and from indoor man-made environments to outdoor natural settings. To standardize evaluation, we will open-source both the training and validation data and the evaluation devkit, facilitating the development and holistic evaluation of more effective inverse graphics algorithms.

We conduct a thorough evaluation of recent methods using our new dataset and existing real and synthetic datasets and provide an analysis of common failure modes. We also explore the effectiveness and limitations of existing datasets and discuss potential improvements. In addition, we put forth a simple baseline integrating a readily available NeuS-based reconstruction~\cite{wangNeuSLearningNeural2021} with an off-the-shelf Mitsuba-based differentiable inverse renderer~\cite{Mitsuba3}. Interestingly, this straightforward approach performs favorably when compared to other methods, including the state-of-the-art relightable neural radiance~\cite{Jin2023TensoIR, zhangNeRFactorNeuralFactorization2021} and neural surface approaches~\cite{hasselgren2022nvdiffrecmc, zhang2022invrender}, across all benchmarks. We conduct several discussions based on our findings with the hope to inspire future research.

In summary, our contributions are as follows. \textbf{i)} We introduce a novel real-world dataset and benchmark for inverse rendering, captured under natural illumination. Our benchmark enables quantitative evaluation on relighting, unlike existing datasets that are either synthetic, qualitative, or confined to lab environments. \textbf{ii)} We propose a straightforward baseline that outperforms existing state-of-the-art methods in relighting and reconstruction across all benchmarks. \textbf{iii)} We provide a systematic evaluation of current state-of-the-art inverse rendering methods and deliver a comprehensive analysis. Our experiments underline the significance of geometry estimation and the use of realistic shading.

%% file: related_work_v2.tex
\section{Related work}
\label{sec:related}

\mypara{Material appearance modeling.}
The Bidirectional Reflectance Distribution Function (BRDF) defines how light is reflected at a surface point, a concept that has been widely used in computer graphics and computer vision to represent surface appearance~\cite{Nicodemus1977}. Many analytical models were proposed, but they often lacked strict physical underpinnings~\cite{blinn1977models, phong1975illumination, ward1992measuring}. This changed with the advent of physically-based reflectance models~\cite{Cook1982, burley2012physically, karis2013,lazarov2013,gotanda2012}, which incorporated physical principles such as energy conservation, leading to their widespread use in rendering engines~\cite{Mitsuba3, blender, freepbr} and inverse rendering methods~\cite{schmitt2020joint, zhangNeRFactorNeuralFactorization2021, huDeepBRDFDeepRepresentation2020}. However, no analytical model can handle all real-world material.

To surpass analytical BRDF limitations, researchers turned to data-driven representations~\cite{matusik2003data, nishino2009directional, kautz2002fast, lombardi2015reflectance, zhangPhySGInverseRendering2021}, with various parameterizations.
Classic representations use dimensionality reduction on lab-measured BRDFs~\cite{matusik2003data}. 
Other low-dimensional parametric representations include mixtures of spherical harmonics~\cite{Wizadwongsa2021NeX,yuPlenoxelsRadianceFields2021, yuPlenOctreesRealTimeRendering2021} and spherical Gaussians~\cite{zhangPhySGInverseRendering2021}.
More recently, deep networks have been used for reflectance modeling~\cite{zhangNeRFactorNeuralFactorization2021, bossNeuralPILNeuralPreIntegrated2021, huDeepBRDFDeepRepresentation2020, sztrajmanNeuralBRDFRepresentation2021, chenInvertibleNeuralBRDF2020}, shown to be more capable than low-dimensional parametric models.

\mypara{Inverse rendering.}
BRDF estimation from images is inherently challenging due to the entanglement of scene lighting and material appearance. Researchers have sought to address this through additional constraints on the capturing setup~\cite{nishino2001determining, matusik2003data, asselin2020deep, Boss2018, Lensch2003, sato1997object, guo2019relightables, biNeuralReflectanceFields2020, nerv2021}. Coded light~\cite{schmitt2020joint} and flashlights have provided valuable cues for estimating specularity~\cite{Nam2018, sang2020single, bi2020deep,biNeuralReflectanceFields2020,biDeep3DCapture2020, Zhang2020InverseRendering, kaya2021uncalibrated}. A number of recent works have sought to estimate BRDFs outside of lab environments by restricting the geometry of surfaces to planes and leveraging the built-in flash of casual cameras~\cite{Aittala2018, Deschaintre2018, henzler2021, Li2018, sang2020single, Deschaintre2020}. Another line of work has recorded videos of rotating objects to estimate reflectance under natural illumination~\cite{dong2014appearance, xia2016recovering}.

Relightable neural radiance field methods, provide another approach to BRDF estimation under natural illumination~\cite{zhangNeRFactorNeuralFactorization2021, bossNeuralPILNeuralPreIntegrated2021, zhangPhySGInverseRendering2021, bossNeRDNeuralReflectance2021, zhang2022invrender, Jin2023TensoIR, neroic}. These methods factorize observed view-conditioned radiance into components corresponding to illumination and materials, offering excellent rendering and relighting quality and promising inverse graphics capabilities under natural environments.

\mypara{Inverse rendering benchmarks.}
Obtaining large-scale, ground-truth data for surface reflectance materials and global illumination in natural environments has thus far proven to be difficult, if not impossible. As a result, many researchers have focused their efforts on qualitative comparisons on surface material estimation or relighting~\cite{barron2015shape}, which unfortunately hinders fair, quantitative comparisons. An alternative approach involves the direct measurement of surface reflectance within a lab environment~\cite{matusik2003data, dupuy2018adaptive, grosse2009groundtruth, shi2019benchmark, li2020multiview}. While these techniques yield accurate BRDF measurements, they often carry limitations related to geometry (such as planar objects), illumination (such as point light), or scene types (usually fixed). 

As photorealistic physics-based rendering has become accessible, synthetic datasets for inverse rendering have surfaced~\cite{li2018interiornet, roberts2021hypersim, zhang2022invrender, liu2023nero, yeh2022photoscene}. These datasets and benchmarks can cover various objects and scenes with ground truth across all attributes. However, assumptions about materials, illumination, and shading must be made during rendering. Furthermore, these datasets still face challenges bridging the simulation-to-reality gap, thus rendering their efficacy in assessing real-world performance questionable.

Researchers have also explored benchmarks that assess downstream tasks like novel view synthesis, relighting, and insertion rendering, which indirectly gauge the quality of inverse rendering. Existing relighting datasets, however, are typically either scene-level (for instance, time-lapses of street scenes)~\cite{rudnev2022nerfosr} or dependent on simple illumination assumptions (such as point lights)~\cite{toschi2023rene,openillumination}. Moreover, some lack proper photometric calibration~\cite{bossNeRDNeuralReflectance2021}, which hinders accurate quantitative measurement. In contrast, our dataset provides a novel benchmark, captured in various real-world environments across multiple viewpoints and illuminations, and offers precise geometric and photometric calibration.

%% file: dataset.tex
\section{Objects with Lighting Dataset}
\label{sec:dataset}

\begin{table*}[ht]
\centering
\begin{adjustbox}{max width=\textwidth}
\begin{tabular}{lcccccc}
\toprule
\bfseries {Dataset} & \bfseries Type & \bfseries {Multi-view} & \bfseries {\#Scenes} & \bfseries {Real data} & \bfseries {Lighting} & \bfseries {\#Envmaps} \\
\midrule
  MIIW \cite{MurmannGAD19}                    & Room (Indoor)      & No & >1000               & Yes          & Limited (Flash)     & - \\
  DiLiGenT \cite{shi2019benchmark}            & Objects (Lab)      & No & 10               & Yes          & Limited (Point light)     & - \\
  DiLiGenT-MV \cite{li2020multiview}          & Objects (Lab)      & Yes & 5               & Yes          & Limited (Point light)     & - \\
  DTU \cite{jensen2014large}                  & Objects (Lab)      & Yes & 128                & Yes          & Limited (LED w/ 7 patterns) & - \\
  Nerf-OSR \cite{rudnev2022nerfosr}           & Buildings (Outdoor) & Yes & 8                  & Yes          & Environment map                   & 44 \\
  OpenIllumination \cite{openillumination}    & Objects (Lab)      & Yes & 64                & Yes          & Limited (Point light patterns) & - \\
  ReNe \cite{toschi2023rene}                  & Objects (Lab)          & Yes & 20                 & Yes          & Limited (Point light)             & - \\
  Glossy-Blender \cite{liu2023nero}           & Glossy objects (In- \& Outdoor) & Yes & 8                  &  No          & Environment map & 6 \\
  Synthetic4Relight \cite{zhang2022invrender} & Objects (In- \& Outdoor)         & Yes & 4                  &  No          & Environment map & 3 \\
  Ours                                        & Objects (In- \& Outdoor)         & Yes & 8                  & Yes          & Environment map & 72 \\
\bottomrule
\end{tabular}
\end{adjustbox}
\caption{Related datasets featuring multiple illuminations. Only our dataset enables multiview reconstruction of real-world objects, relighting with measured complex environment maps, and ground truth data for quantitative evaluation.}
\label{tab:datasets_overview}
\end{table*}

\begin{figure*}
    \centering
    \begin{adjustbox}{width=\textwidth}
    \adjustimage{rotate=3, height=2.7cm}{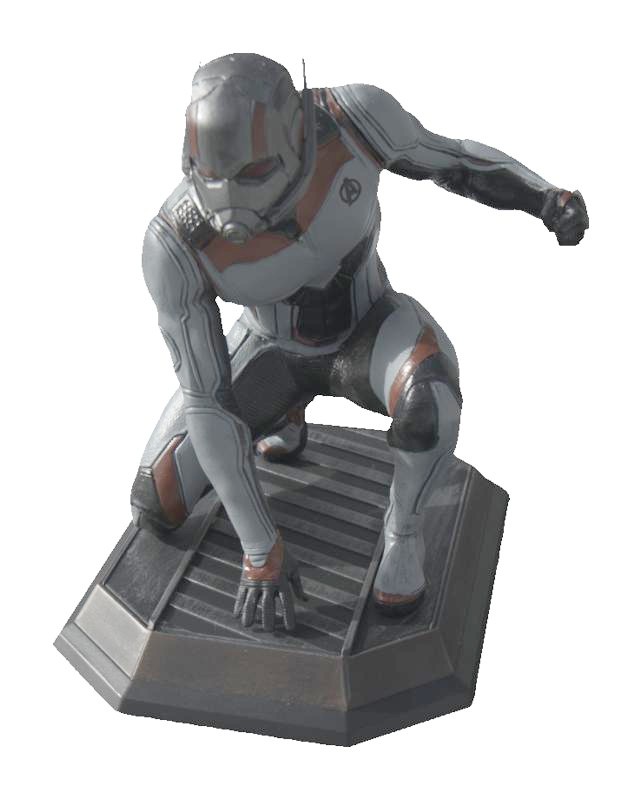}%
    \adjustimage{height=2cm, margin=2mm 0mm 2mm 0mm}{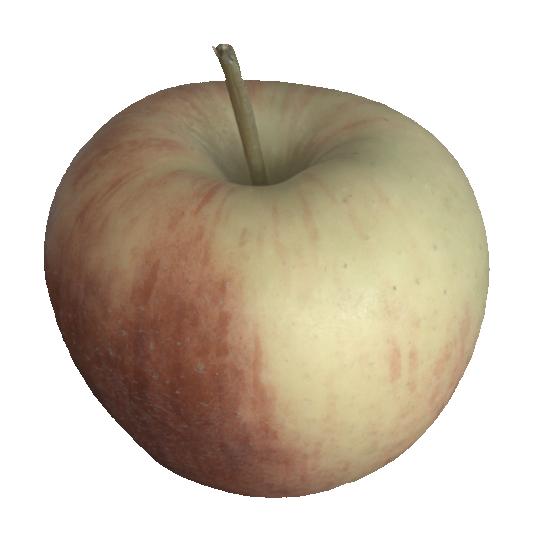}%
    \adjustimage{height=3cm}{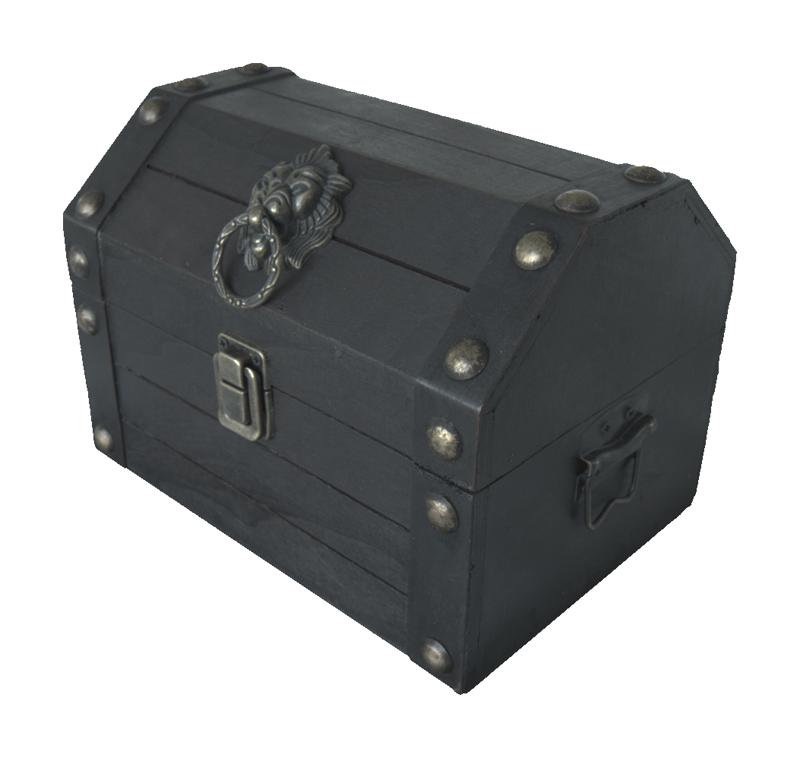}%
    \adjustimage{height=1.9cm}{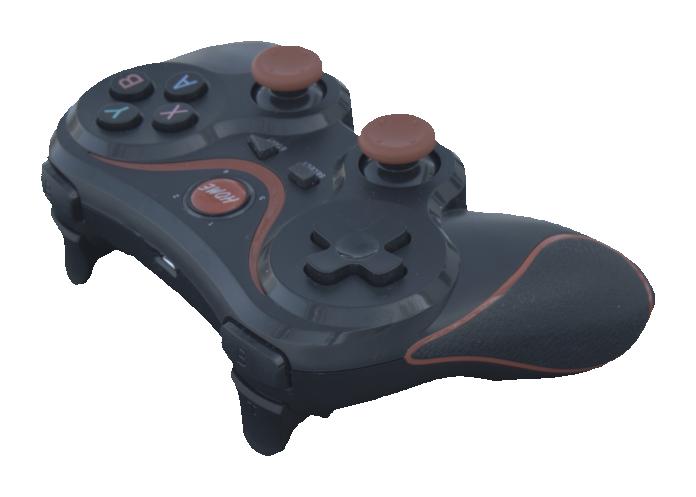}%
    \adjustimage{height=3.8cm}{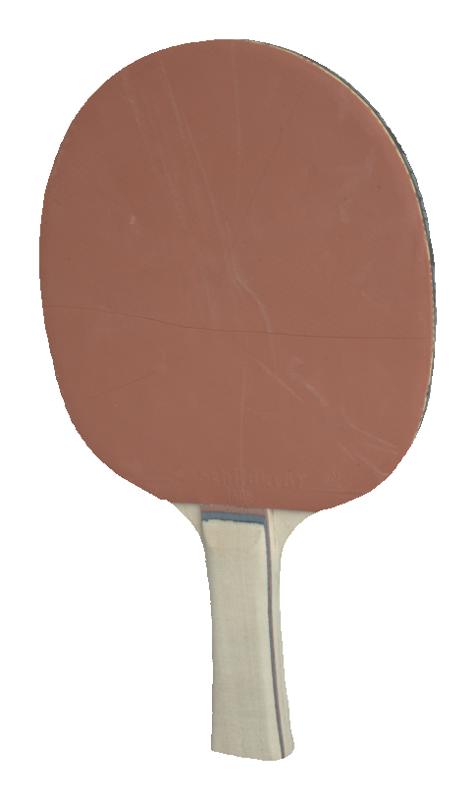}%
    \adjustimage{height=2cm, margin=2mm 0mm 2mm 0mm}{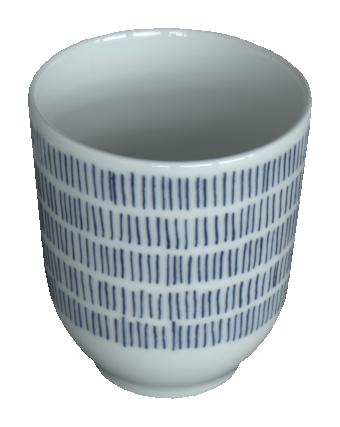}%
    \adjustimage{height=2.7cm, margin=2mm 0mm 2mm 0mm}{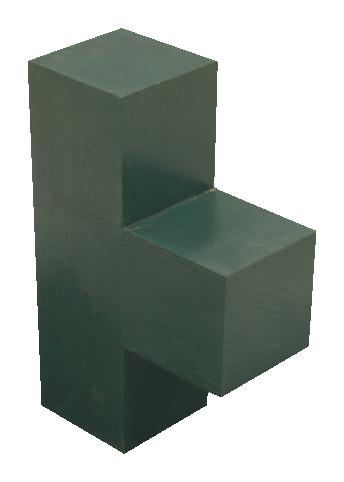}%
    \adjustimage{height=2.8cm}{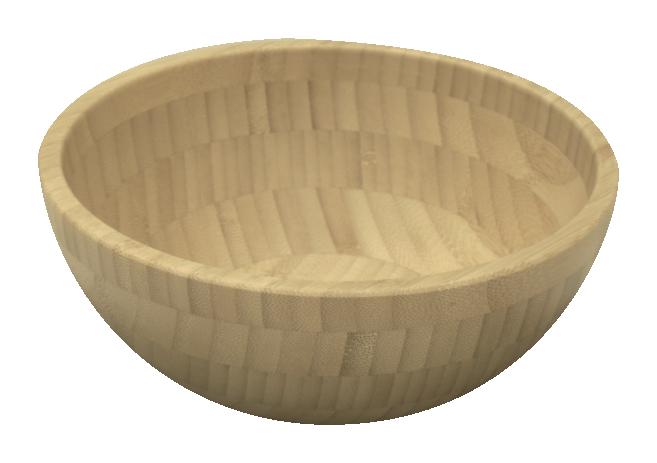}%
    \end{adjustbox}\\[2mm]%
    \begin{adjustbox}{width=\textwidth}
    \adjustimage{height=2cm}{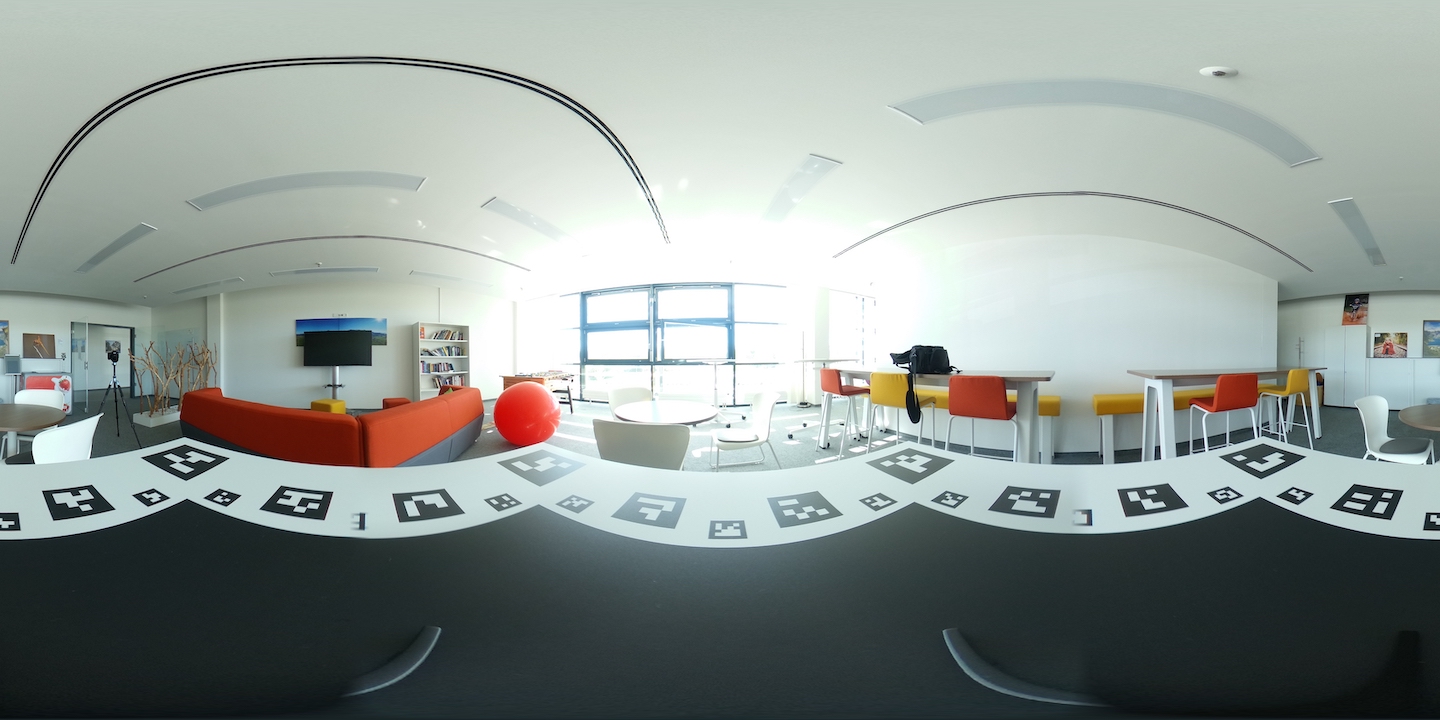}%
    \adjustimage{height=2cm}{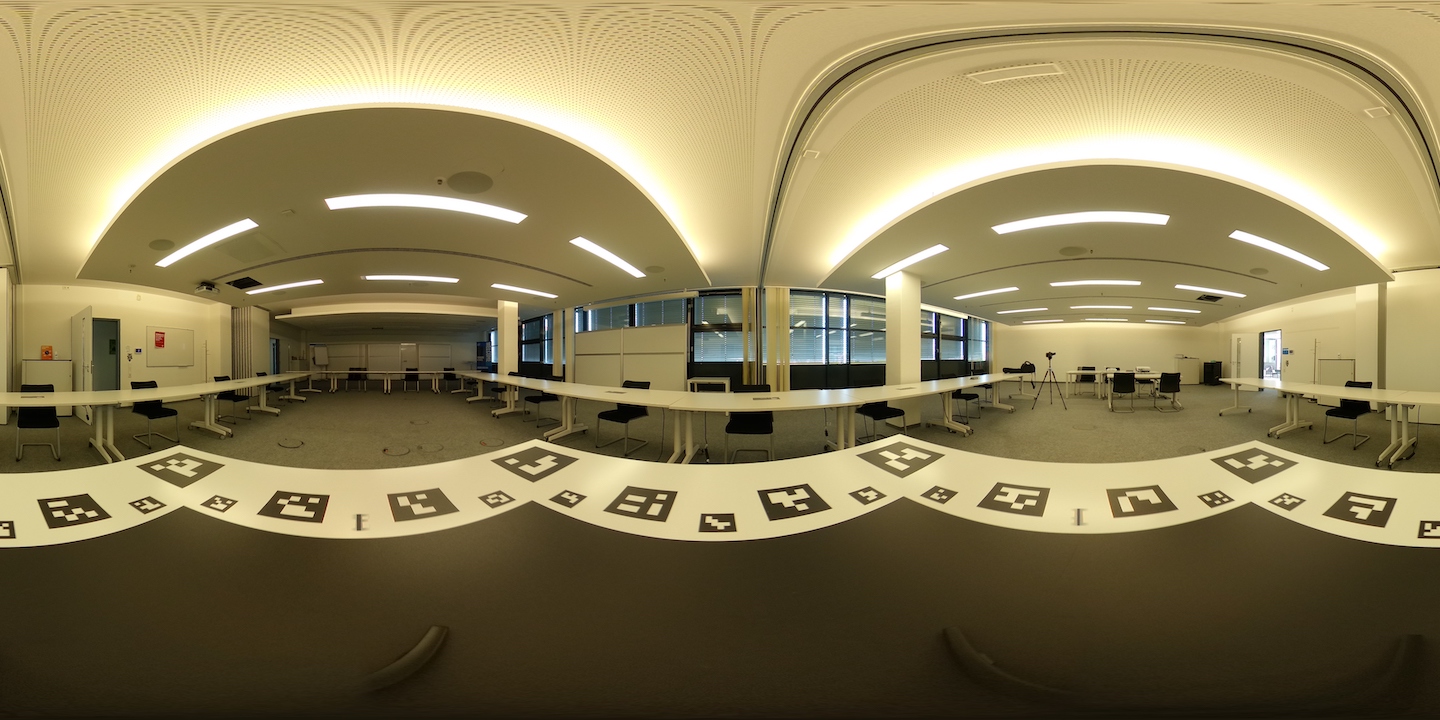}%
    \adjustimage{height=2cm}{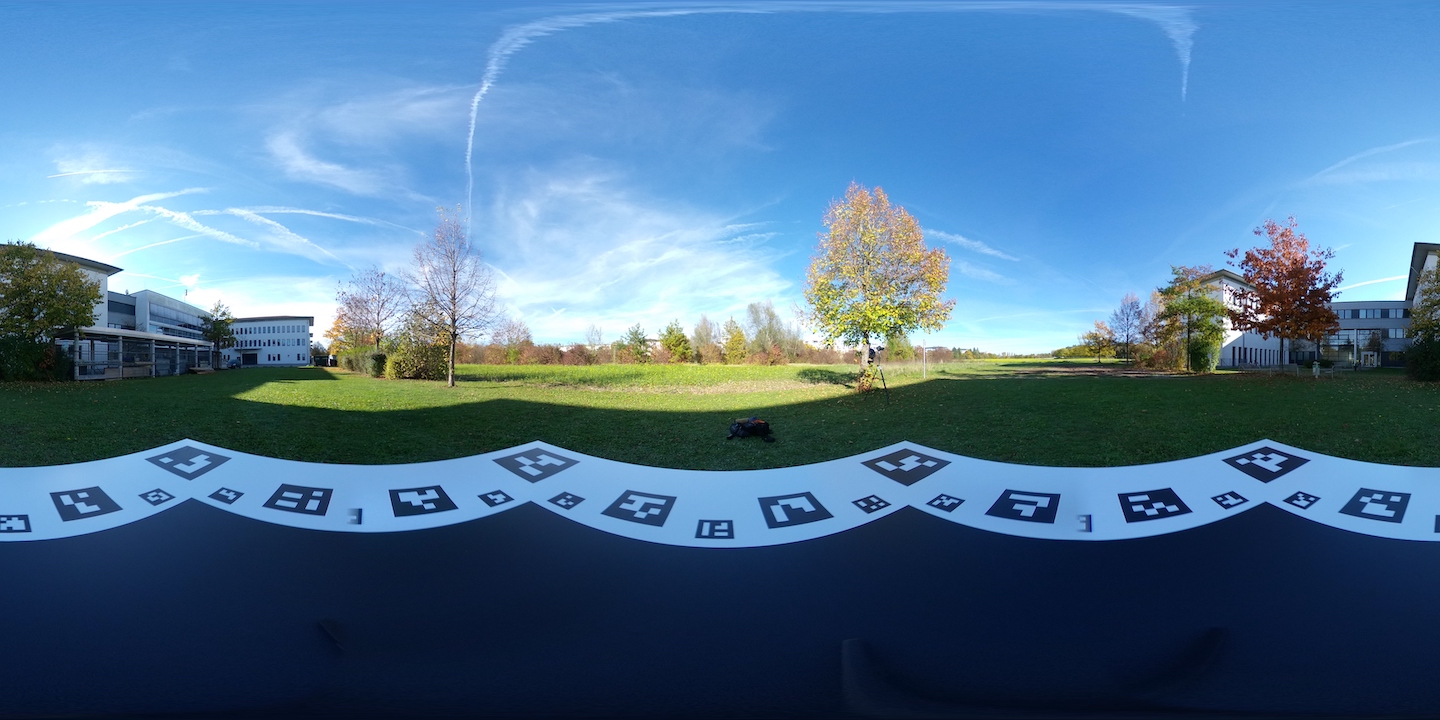}%
    \end{adjustbox}\\[2mm]%
    \begin{adjustbox}{width=\textwidth}
    \adjustimage{height=2cm}{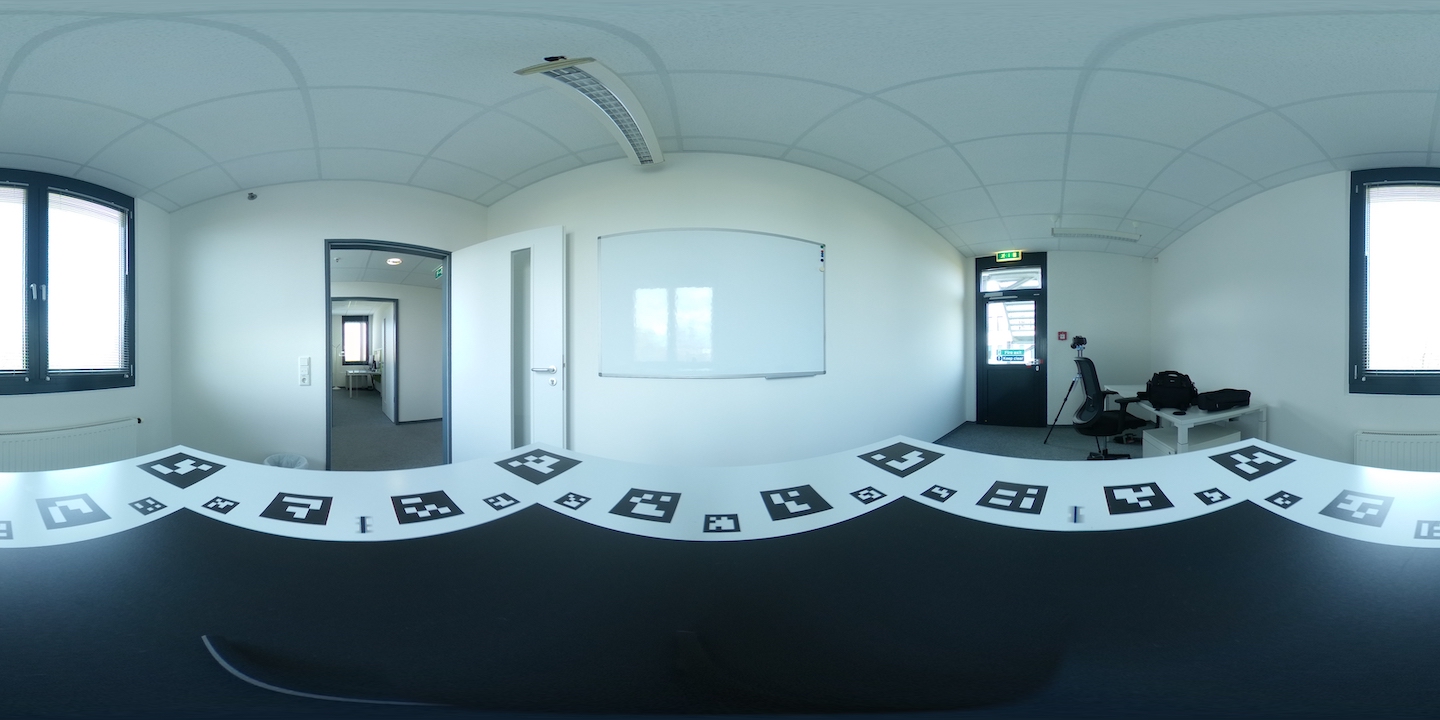}%
    \adjustimage{height=2cm}{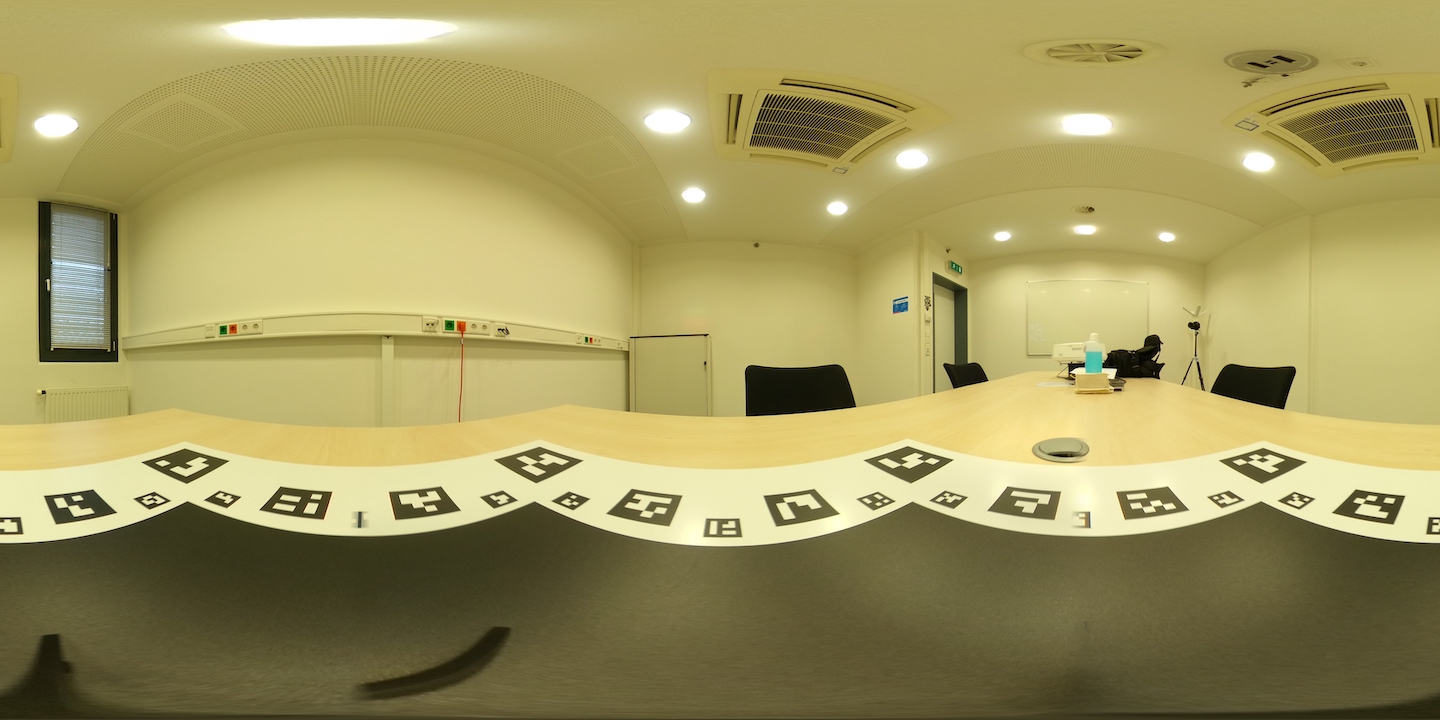}%
    \adjustimage{height=2cm}{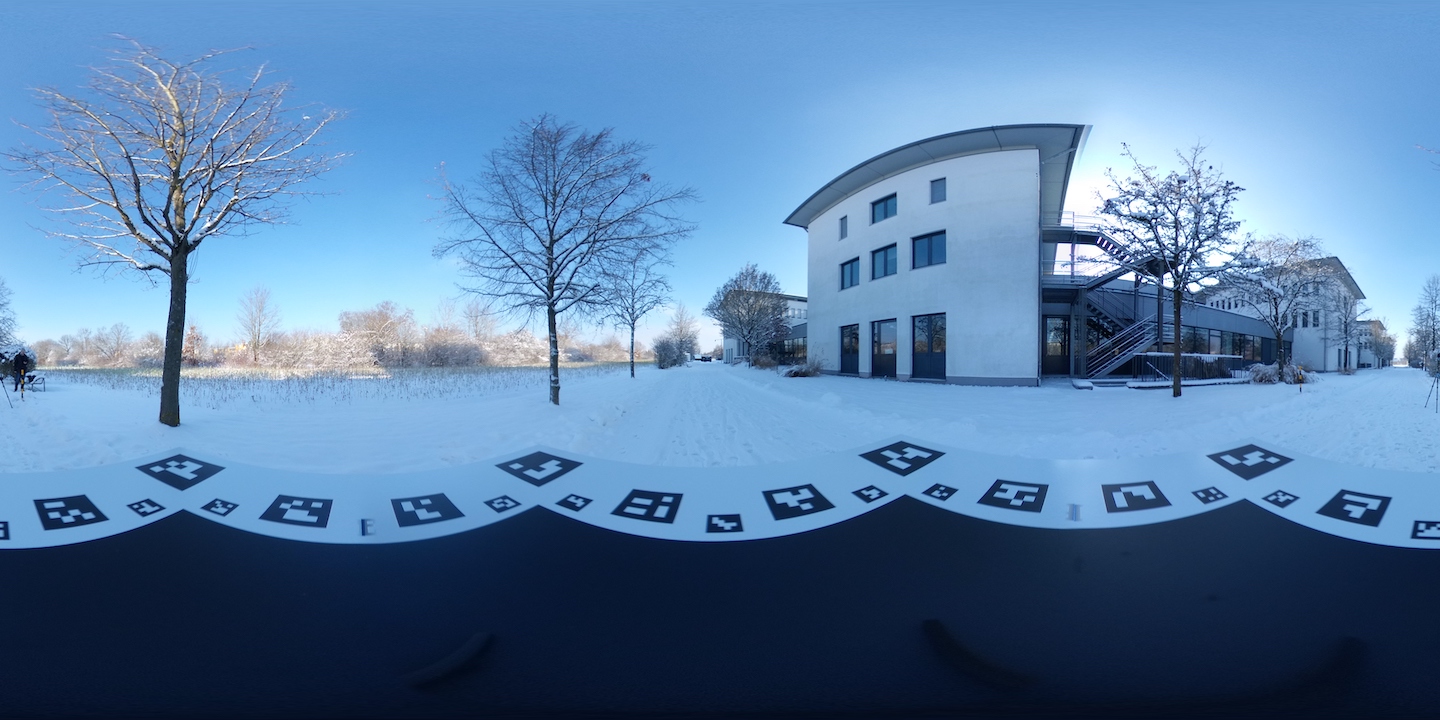}
    \end{adjustbox}
    \caption{Overview of the Objects With Lighting dataset. \textbf{First row:} Our dataset contains 8 objects which we capture in 3 different lighting conditions each. \textbf{Second and third row:} We show 2 HDR environment map examples for each lighting category which are (left to right): indoor with natural light, indoor with artificial light, and outdoor.}
    \label{fig:dataset}
\end{figure*}

We present the Objects With Lighting dataset~\footnote{The Objects With Lighting dataset is released under the CDLA-permissive 2.0 license.}, an open dataset specifically designed around faithful relighting of real-world objects in new environments.
The Objects With Lighting dataset is more suited for this task than other related datasets shown in \tab{tab:datasets_overview}.
Current datasets present limitations: they are either synthetic, provide only simplified lighting conditions, or just do not suit the problem, as they lack multiple views or involve specialized gear.

Our dataset features eight objects, each captured in three distinct environments with intricate lighting. To ensure varied lighting conditions, we have chosen these environments to include outdoor, indoor, and indoor settings with artificial light.
\fig{fig:dataset} shows surface reconstructions of these objects and a subset of the environments.
Images and environment maps have been geometrically calibrated and aligned to a common color space to enable quantitative evaluation of rendered relighted objects against the ground truth images.
In each environment, we have 3 test images, each with its unique environment map. This map is designed to account for the specific camera that captured the ground truth image.
For evaluation, we use expert annotated masks to select the object pixels in the ground truth images.
For geometric calibration, we combine natural features using SIFT descriptors, fiducial markers, and expert keypoint annotation to ensure high-quality camera poses.
To capture the environment lighting we create unclipped HDRI using a Theta Z1 360-degree camera.
For the input images and ground truth, we capture images at high resolution using a Canon EOS 90D DSLR camera.
We use RAW format for all images and implement our own image processing pipeline to control all transformations of the source material.
For color calibration, we use a 24-color chart and transform the images from 
both cameras %
to a common color space. For more details on the calibration process, please refer to \sectn{sec:data-collection} in the appendix.

\begin{figure}
    \centering
    \adjustimage{width=0.50\textwidth}{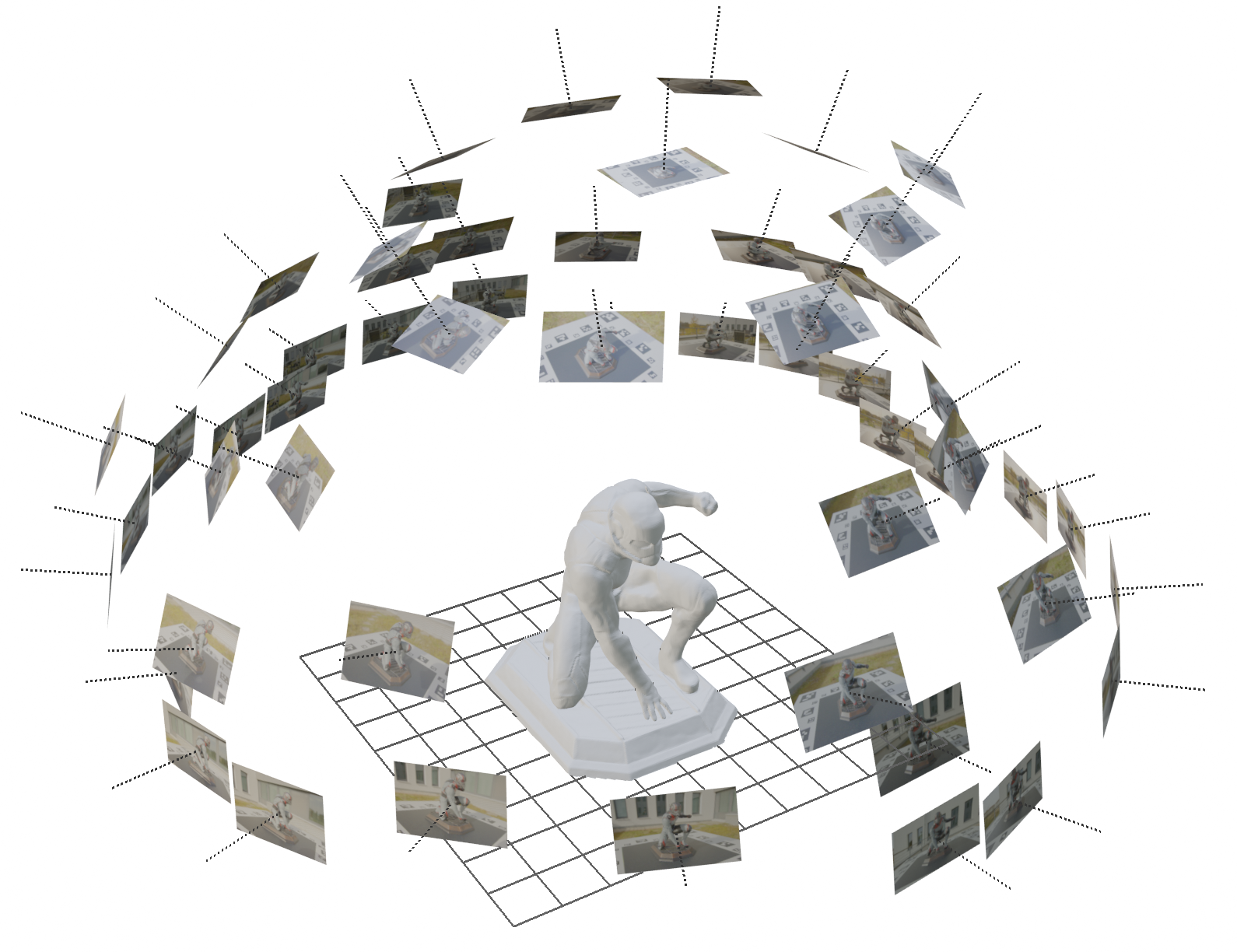}%
    \caption{We take 42-67 photos for each object and roughly sample a hemisphere with the object in the center.}
    \label{fig:cameraposes}
\end{figure}

A key motivation behind our dataset is to assist with the material reconstruction tasks. To this end, we seek to facilitate the surface reconstruction task through a comprehensive object coverage, by sampling images on a hemisphere, as depicted in~\fig{fig:cameraposes} . We select objects made from diverse materials like plastic, wood, metal, porcelain, rubber, and fruit skin. These materials are not uniform but spatially vary, posing a challenge to reconstruction methods.
In addition to the core data files comprised of posed images, we provide additional resources for convenience. These include approximate foreground masks for input images, and NeuS~\cite{wangNeuSLearningNeural2021} generated meshes, which can be used to simplify or skip shape reconstruction for methods focusing on material estimation or fine-tuning. We leverage this additional data to compare multiple approaches in our experiments.

%% file: method.tex
\section{A simple baseline}
\label{sec:method}

Our baseline method serves as a check to see if we can solve the problem with existing easy to use building blocks.
To keep the method simple, we choose a sequential approach, which uses NeuS \cite{wangNeuSLearningNeural2021} to reconstruct the shape of the objects first. 
NeuS optimizes a function $f(\xx)$ that maps a 3D point $\xx$ to a scalar signed distance value and a function $c(\xx,\vv)$ which maps a position $\xx$ and direction $\vv$ to a color. 
Both functions are implemented as MLPs and differentiable volume rendering is used to optimize the network parameters.
After optimization the surface $S$, can be extracted from $f$ using the marching cubes algorithm \cite{lorensen1987marching}.
We refer to \cite{wangNeuSLearningNeural2021} for a detailed description of the algorithm.

In a second step we compute the material parameters via differentiable rendering with the path tracer Mitsuba \cite{Mitsuba3}.
The loss we optimize is 
\begin{equation}
\begin{aligned}
    \lL = & \sum_i^N \int \Vert I_i - \hat I_i \Vert \mathrm d\xx + \alpha_1 \int \Vert \nabla \theta_\text{env} \Vert \mathrm d\xx \\
          & + \alpha_2 \left( \int \Vert \nabla \theta_\text{albedo} \Vert + \Vert \nabla \theta_\text{rough} \Vert + \Vert \nabla \theta_\text{metal}\Vert \mathrm d\xx \right) .
    \label{eq:baseline_loss}
\end{aligned}
\end{equation}
We minimize the difference between the input images $I_i(\xx)$ and the rendered images $\hat I_i(\xx)$ for all pixels $\xx$.
$\hat I_i$ is the rendered image that depends on the mesh $S$ obtained from NeuS, the camera parameters, the material parameters $\theta = \{\theta_\text{albedo}, \theta_\text{rough}, \theta_\text{metal}\}$ of the principled BRDF \cite{burley2012physically}, and the environment map $\theta_\text{env}$. 
We keep all other parameters of the BRDF model constant at their default values.
We store the material parameters and environment map as 2D textures to allow them to spatially vary over the surface and sphere and regularize them with a simple total variation regularization weighted with $\alpha_1 = \alpha_2 =0.01$.

The rendering process that we use is governed by 
\begin{equation}
    L_o(\bomega_o) = \int_\Omega L_i(\bomega_i)\,f(\bomega_i,\bomega_o, \theta)\,\langle\bn\,,\bomega_i\rangle\; \mathrm d\bomega_i,
    \label{eq:rendering}
\end{equation}
which is known as the rendering equation \cite{kajiya1986} and computed using Monte Carlo integration over the hemisphere $\Omega$ for a point on the surface with normal $\nn$.
$L_i$ and $L_o$ are the ingoing and outgoing radiance along the directions $\bomega_i, \bomega_o$.
The parameters we want to optimize are part of the BRDF $f$ and $L_i$ which is directly and indirectly affected by $\theta_\text{env}$.
We refer to \cite{Mitsuba3} for details on how the rendering equation can be implemented and its parameters optimized efficiently.

%% file: experiments.tex
\section{Experiments}
\label{sec:experiments}

We conduct experiments on our and related datasets with our baseline method and recent state-of-the-art methods.
We present quantitative evaluations for relighting in comparison with a synthetic relighting dataset and discuss failure modes.
Following common practices we measure novel view synthesis performance on multiple real and synthetic datasets too and check for correlations with the more challenging relighting task.

\subsection{Datasets}
We use four datasets for our experiments.
(i) Our \mbox{Objects With Lighting} dataset introduced in \sectn{sec:dataset}. %
(ii) The Synthetic4Relight dataset from \cite{zhang2022invrender}, which features synthetic images and ground truth for relighting experiments.
(iii) A subset of DTU \cite{jensen2014large} which is a real-world dataset for evaluating Multi-View Stereo methods.
(iv) A subset of the synthetic BlendedMVS (BMVS) dataset \cite{yao2020blendedmvs}.%
We use DTU and BMVS as additional datasets for evaluating novel view synthesis performance.

\subsection{Test protocol}
\begin{figure*}
    \centering
    \adjustimage{width=1.0\textwidth}{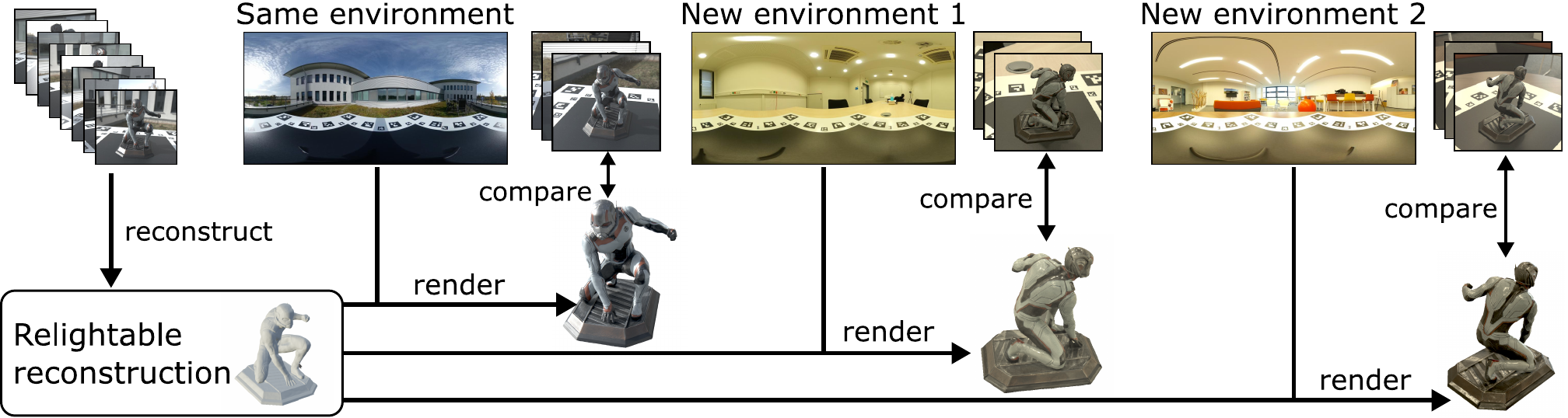}
    \caption{Test protocol on the \emph{Objects With Lighting} dataset. 
    We reconstruct a relightable representation of the object from a set of images taken in one environment. The reconstruction can be a textured mesh, a neural representation, or any other representation that can be relit with a new environment map and rendered from a novel view point. 
    Testing is conducted by rendering the objects using environment maps from the reconstruction environment and new environments and comparing the rendered images to the corresponding test images of that environment.}
    \label{fig:test-protocol}
\end{figure*}

\mypara{Relighting} The test protocol for relighting is visualized in \fig{fig:test-protocol}.
For each method, we reconstruct the object generating a method-specific representation from a set of images taken in the reconstruction environment.
After reconstruction, we render the object under new lighting using an environment map and new viewpoints that correspond to the ground truth test images.
Directly testing the rendered images allows us to evaluate all methods uniformly irrespective of the used representations.
When testing, one of the environment maps corresponds to the reconstruction environment but is not available during the reconstruction process.
Results for this environment are listed as \emph{Same environment} in Tables~\ref{tab:object-relighting} and \ref{tab:synth4relight}.

To account for the scalar ambiguity between illumination and albedo of the surface we postprocess the rendered linear images and compute the optimal exposure for minimizing the MSE between the rendered tone mapped image and the ground truth image for each channel.
For tone mapping and conversion to 8-bit images we use $y = 255 (2^\text{EV} x)^\gamma$.
Values $x$ are linear values and $y$ are the tone mapped 8-bit values.
We round all values to the nearest integer and clip values outside the 8-bit range.
$\gamma = \frac{1}{2.2}$ for all experiments.
To measure performance we use the standard metrics PSNR, SSIM, and LPIPS \cite{zhang2018perceptual} to compare images.
We compute all metrics per-pixel and exclude background pixels from the averages using ground truth masks to avoid very large or very small values for the respective metrics.
We use the same evaluation code for our dataset and for Synthetic4Relight.

\mypara{Novel view synthesis}
For novel view synthesis we use the lighting estimated by each method instead of the measured environment maps for rendering. %
Another difference to relighting is the tone mapping.
Since we use the estimated lighting we do not have an unknown scalar factor that relates the material and the illumination.
Thus, we skip any post-processing and compare tone mapped images as generated by each method.
We compute the metrics with the same code as for the relighting experiments.

\subsection{Methods}
We evaluate several state-of-the-art methods and our baseline described in \sectn{sec:method}, which we name \emph{Mitsuba+NeuS}.
Common to all methods is the use of differential rendering to optimize shape and appearance.
Methods differ in the way they represent shape, materials, and lighting.

NVDiffrec \cite{munkbergExtractingTriangular3D2021} and NVDiffrecMC \cite{hasselgren2022nvdiffrecmc} use tetrahedral grids to represent shapes during reconstruction.
The remaining methods use implicit neural representations to describe the surface.
Materials represented with analytical BRDF models is used by our baseline, NVDiffrec, NVDiffrecMC, PhySG \cite{zhangPhySGInverseRendering2021}, NeRD \cite{bossNeRDNeuralReflectance2021} and NeROIC \cite{neroic}.
InvRender \cite{Zhang2020InverseRendering} and NeuralPIL \cite{bossNeuralPILNeuralPreIntegrated2021} make use of analytical BRDFs but combine them with neural representations to enforce sparsity and improve convergence.
TensoIR \cite{Jin2023TensoIR} combines neural rendering and physically-based rendering and uses joint optimization.
For novel view synthesis experiments we use the neural rendering capabilities while we use the physically-based rendering for the relighting experiments.
NeRFactor \cite{zhangNeRFactorNeuralFactorization2021} uses albedo and a latent BRDF code trained on a material database to represent materials.
Illuminations is represented as equirectangular environment map in NVDiffrec, NVDiffrecMC, TensoIR, NeRFactor, and our baseline.
InvRender, PhySG, and NeRD use spherical Gaussians.
NeROIC makes use of spherical harmonics and Neural-PIL uses a neural representation to store preintegrated illumination.

\subsection{Relighting}
\input{images_new_owl_table}
\input{images_new_synth4relight_table}

We evaluate all methods on our \emph{Objects With Lighting} dataset and report results in \tab{tab:object-relighting}.
We found that it is challenging for many methods to compute a good shape representation on our and other datasets, which can result in failures producing very large errors.
To alleviate this we allow the usage of approximate foreground masks or inject the meshes generated with NeuS, which we denote with the suffix \emph{+NeuS}.
On the Synthetic4Relight dataset we use foreground masks for all methods and show results in \tab{tab:synth4relight}. 

Results show that relighting with illumination from new environments is more difficult than relighting with the environment map from the reconstruction environment showing the importance of capturing objects in additional environments.

Despite being a simple composition of existing methods our baseline Mitsuba+NeuS performs well on our dataset and the Synthetic4Relight dataset.
On the synthetic dataset we observe very high absolute values for PSNR and SSIM as well as a bigger lead to other methods.
It is striking that NVDiffrecMC+NeuS achieves a very high PSNR on the synthetic data too.
Mitsuba and NVDiffrecMC use both Monte Carlo ray tracing and the same holds for the Cycles renderer, which was used to generate the synthetic data.
Using similar rendering techniques and BRDF representations may produce unwanted resonance between the data and the methods.
This may lead to overly optimistic results and is a problem that can only be resolved by using real-world data.

\begin{figure*}
    \centering
    \adjustimage{width=0.78\textwidth}{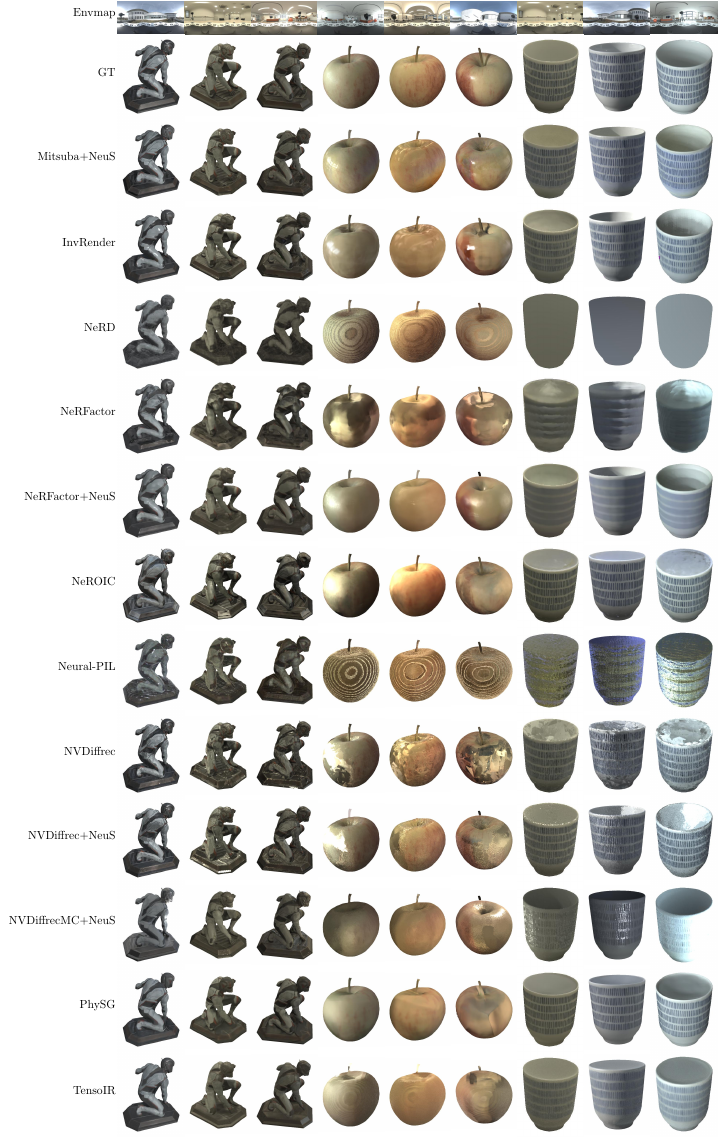}
    \caption{Qualitative examples for relighting on our dataset.
    The first row shows the environment map used for relighting.
    For each object, we show one of the test images of the 3 lighting categories.
    The first image for each object corresponds to the reconstruction environment. Each object has been captured in highly different lighting conditions: outdoor, indoor, and indoor with artificial light.
    }
    \label{fig:ord_examples}
\end{figure*}

Our dataset helps us to identify failure modes.
The most obvious failures are caused by bad shape reconstruction.
Real-world data is more challenging to reconstruct even if foreground masks are available.
We can see this especially for NeRD, NeRFactor, and NVDiffrec in \fig{fig:ord_examples} which fail for the apple and porcelain mug objects.
While capturing the concave shape of the mug is challenging for all methods, NeRFactor, NeROIC, Neural-PIL, NVDiffrec, PhySG, and TensoIR miss to reconstruct the inside completely.

Another deficiency we can observe is the overestimation of the glossyness of objects.
This can be observed well for our baseline, InvRender, and NVDiffrecMC+NeuS in the highlights on the apple but is also visible for the antman object, which shows highlights not visible in the ground truth. %
It can also be seen for the gamepad object for NVDiffrecMC+NeuS and InvRender (see \fig{fig:ord_examplesA} and \fig{fig:ord_examplesB} in the appendix for qualitative examples of all objects).
We did not observe these failures in the Synthetic4Relight experiments.

For NeRD and Neural-PIL we observe artifacts on real data. For Neural-PIL we see similar artifacts on the Synthetic4Relight dataset too but find that the artifacts are more pronounced on real data.

With respect to the rendering we see that methods that don't account for visibility with respect to the illumination have a disadvantage due to missing shadows.
This can be seen best for NVDiffrec+NeuS which misses reproducing the shadow inside the porcelain mug.

\subsection{Novel view synthesis}
\input{images_new_nvs_table_v1b}

Novel view synthesis is a related but simpler problem since we can assume that the illumination is fixed.
We use all datasets for evaluation. %
Note that we can use test images from the reconstruction environment of our dataset to test novel view synthesis performance.

We report results in \tab{tab:nvs}, which shows that our baseline Mitsuba+NeuS performs well on this task too.
We check for correlations between the relighting experiments on our dataset and the novel-view performance across all datasets.
We find a significant correlation when ranking the methods using the LPIPS metric with $r_s=0.77$ and $p=0.02$, and the SSIM metric with $r_s=0.91$ and $p=0.002$.
For the PSNR metrics, we cannot find a significant correlation between the tasks. %
Correlation is expected since we assume that a method that solves the relighting task perfectly will solve the novel view synthesis task too.
However, not seeing a clear correlation for PSNR and the small sample size ($N=9$) do not allow for a final conclusion.
Additionally, we know that novel view synthesis does not easily expose problems like baked-in or missing shadows, something we can observe on our dataset, thus we think results as shown in \tab{tab:nvs} must be interpreted together with quantitative relighting experiments.

%% file: images_new_owl_table.tex
\begin{table*}
\renewcommand{\thefootnote}{\fnsymbol{footnote}}
\centering
\footnotesize
\begin{tabular}{lS[table-format=2.2,detect-all]S[table-format=1.2,detect-all]S[table-format=1.3,detect-all]S[table-format=2.2,detect-all]S[table-format=1.2,detect-all]S[table-format=1.3,detect-all]S[table-format=2.2,detect-all]S[table-format=1.2,detect-all]S[table-format=1.3,detect-all]}
\toprule
                   & \multicolumn{3}{c}{\bfseries {Same environment}}             & \multicolumn{3}{c}{\bfseries {New environment}}              & \multicolumn{3}{c}{\bfseries {Mean}}                         \\ \cmidrule(lr){2-4} \cmidrule(lr){5-7} \cmidrule(lr){8-10}
\bfseries {Method} & {PSNR $\uparrow$} & {SSIM $\uparrow$} & {LPIPS $\downarrow$} & {PSNR $\uparrow$} & {SSIM $\uparrow$} & {LPIPS $\downarrow$} & {PSNR $\uparrow$} & {SSIM $\uparrow$} & {LPIPS $\downarrow$} \\ \midrule
{Mitsuba+NeuS}     & \bfseries 26.20   & \bfseries 0.86    & \bfseries 0.205      & \bfseries 26.24   & \bfseries 0.84    & \bfseries 0.227      & \bfseries 26.22   & \bfseries 0.84    & \bfseries 0.219      \\
{InvRender\footnotemark[1]}        & 24.42             & 0.78              & 0.360                & 23.45             & 0.77              & 0.374                & 23.77             & 0.78              & 0.369                \\
{NeRD\footnotemark[1]}             & 22.68             & 0.63              & 0.554                & 21.71             & 0.65              & 0.540                & 22.04             & 0.64              & 0.544                \\
{NeRFactor\footnotemark[1]}        & 22.15             & 0.75              & 0.452                & 20.62             & 0.72              & 0.486                & 21.13             & 0.73              & 0.475                \\
{NeRFactor+NeuS}   & 25.22             & 0.82              & 0.393                & 25.82             & 0.81              & 0.411                & 25.62             & 0.81              & 0.405                \\
{NeROIC\footnotemark[1]}           & 22.81             & 0.82              & 0.305                & 21.59             & 0.78              & 0.323                & 22.00             & 0.79              & 0.317                \\
{Neural-PIL\footnotemark[1]}       & 19.88             & 0.50              & 0.602                & 19.56             & 0.51              & 0.604                & 19.67             & 0.51              & 0.604                \\
{NVDiffrec\footnotemark[1]}        & 24.45             & 0.79              & 0.332                & 22.60             & 0.72              & 0.406                & 23.21             & 0.74              & 0.381                \\
{NVDiffrec+NeuS}   & 24.47             & 0.82              & 0.286                & 22.78             & 0.74              & 0.373                & 23.34             & 0.77              & 0.344                \\
{NVDiffrecMC\footnotemark[1]}      & 18.99             & 0.74              & 0.379                & 20.24             & 0.73              & 0.393                & 19.82             & 0.73              & 0.389                \\
{NVDiffrecMC+NeuS} & 19.14             & 0.76              & 0.346                & 20.32             & 0.74              & 0.371                & 19.92             & 0.75              & 0.362                \\
{PhySG\footnotemark[1]}            & 23.49             & 0.82              & 0.360                & 22.77             & 0.82              & 0.375                & 23.01             & 0.82              & 0.370                \\
{TensoIR\footnotemark[1]}          & 25.60             & 0.80              & 0.348                & 24.15             & 0.77              & 0.378                & 24.64             & 0.78              & 0.368                \\
\bottomrule
\end{tabular}
\caption{Relighting results on our \emph{Objects With Lighting} dataset. \protect\footnotemark[1]{Uses foreground segmentation masks}.}
\label{tab:object-relighting}
\end{table*}

%% file: images_new_synth4relight_table.tex
\begin{table*}
\centering
\footnotesize
\begin{tabular}{lS[table-format=2.2,detect-all]S[table-format=1.2,detect-all]S[table-format=1.3,detect-all]S[table-format=2.2,detect-all]S[table-format=1.2,detect-all]S[table-format=1.3,detect-all]S[table-format=2.2,detect-all]S[table-format=1.2,detect-all]S[table-format=1.3,detect-all]}
\toprule
                    & \multicolumn{3}{c}{\bfseries {Same environment}}             & \multicolumn{3}{c}{\bfseries {New environment}}              & \multicolumn{3}{c}{\bfseries {Mean}}                         \\ \cmidrule(lr){2-4} \cmidrule(lr){5-7} \cmidrule(lr){8-10}
\bfseries {Method} & {PSNR $\uparrow$} & {SSIM $\uparrow$} & {LPIPS $\downarrow$} & {PSNR $\uparrow$} & {SSIM $\uparrow$} & {LPIPS $\downarrow$} & {PSNR $\uparrow$} & {SSIM $\uparrow$} & {LPIPS $\downarrow$} \\ \midrule
{Mitsuba+NeuS}     & \bfseries 29.02   & \bfseries 0.88    & \bfseries 0.106      & \bfseries 30.47   & \bfseries 0.88    & \bfseries 0.108      & \bfseries 29.99   & \bfseries 0.88    & \bfseries 0.107      \\
{InvRender}        & 22.08             & 0.76              & 0.297                & 21.15             & 0.77              & 0.279                & 21.46             & 0.77              & 0.285                \\
{NeRD}             & 22.15             & 0.61              & 0.424                & 22.95             & 0.65              & 0.356                & 22.68             & 0.64              & 0.378                \\
{NeRFactor}        & 23.30             & 0.69              & 0.367                & 21.60             & 0.62              & 0.341                & 22.17             & 0.65              & 0.350                \\
{NeROIC}           & 19.07             & 0.68              & 0.358                & 18.35             & 0.66              & 0.338                & 18.59             & 0.66              & 0.345                \\
{Neural-PIL}       & 16.68             & 0.45              & 0.480                & 17.37             & 0.47              & 0.437                & 17.14             & 0.47              & 0.452                \\
{NVDiffrec}        & 22.60             & 0.83              & 0.206                & 21.03             & 0.75              & 0.228                & 21.55             & 0.78              & 0.220                \\
{NVDiffrecMC+NeuS} & 26.80             & 0.81              & 0.185                & 27.65             & 0.82              & 0.177                & 27.36             & 0.81              & 0.180                \\
{PhySG}                 & 21.03             & 0.72              & 0.383                & 21.46             & 0.72              & 0.349                & 21.32             & 0.72              & 0.360                \\
{TensoIR}          & 26.17             & 0.80              & 0.246                & 27.27             & 0.82              & 0.202                & 26.90             & 0.81              & 0.217                \\
\bottomrule
\end{tabular}
\caption{Relighting results on the Synthetic4Relight dataset.}
\label{tab:synth4relight}
\end{table*}

%% file: images_new_nvs_table_v1b.tex
\begin{table*}
\setlength{\tabcolsep}{2pt}
\centering
\begin{adjustbox}{max width=\textwidth}
\begin{tabular}{@{}lS[table-format=2.2,detect-all]S[table-format=1.2,detect-all]S[table-format=1.3,detect-all]S[table-format=2.2,detect-all]S[table-format=1.2,detect-all]S[table-format=1.3,detect-all]S[table-format=2.2,detect-all]S[table-format=1.2,detect-all]S[table-format=1.3,detect-all]S[table-format=2.2,detect-all]S[table-format=1.2,detect-all]S[table-format=1.3,detect-all]S[table-format=2.2,detect-all]S[table-format=1.2,detect-all]S[table-format=1.3,detect-all]@{}}
    \toprule
                        & \multicolumn{3}{c}{\bfseries {DTU}}                          & \multicolumn{3}{c}{\bfseries {BMVS}}                         & \multicolumn{3}{c}{\bfseries {Synthetic4Relight}}            & \multicolumn{3}{c}{\bfseries {Objects With Lighting}}       & \multicolumn{3}{c}{\bfseries {Mean}}                         \\ \cmidrule(lr){2-4} \cmidrule(lr){5-7} \cmidrule(lr){8-10} \cmidrule(lr){11-13} \cmidrule(lr){14-16}
    \bfseries {Method} & {PSNR $\uparrow$} & {SSIM $\uparrow$} & {LPIPS $\downarrow$} & {PSNR $\uparrow$} & {SSIM $\uparrow$} & {LPIPS $\downarrow$} & {PSNR $\uparrow$} & {SSIM $\uparrow$} & {LPIPS $\downarrow$} & {PSNR $\uparrow$}    & {SSIM $\uparrow$} & {LPIPS $\downarrow$} & {PSNR $\uparrow$} & {SSIM $\uparrow$} & {LPIPS $\downarrow$} \\ \midrule
    {Mitsuba+NeuS}     & \bfseries 20.49   & \bfseries 0.65    & \bfseries 0.241      & 22.18             & \bfseries 0.69    & \bfseries 0.225      & \bfseries 29.64   & \bfseries 0.88    & \bfseries 0.091      & 25.08                & 0.87              & \bfseries 0.186      & \bfseries 24.35   & \bfseries 0.77    & \bfseries 0.186      \\
    {InvRender}        & 17.09             & 0.55              & 0.439                & 18.66             & 0.50              & 0.506                & 22.15             & 0.76              & 0.294                & 24.44                & 0.81              & 0.336                & 20.59             & 0.66              & 0.394                \\
    {NeRD}             & 18.85             & 0.49              & 0.434                & 20.17             & 0.46              & 0.489                & 21.37             & 0.62              & 0.408                & 20.52                & 0.66              & 0.535                & 20.23             & 0.56              & 0.467                \\
    {NeRFactor}        & 18.47             & 0.54              & 0.412                & 19.77             & 0.49              & 0.481                & 20.39             & 0.69              & 0.363                & 22.99                & 0.78              & 0.434                & 20.41             & 0.63              & 0.423                \\
    {NeROIC}           & 18.90             & \bfseries 0.65    & 0.357                & 20.37             & 0.61              & 0.367                & 20.04             & 0.73              & 0.332                & 20.18                & 0.82              & 0.322                & 19.87             & 0.70              & 0.345                \\
    {Neural-PIL}       & 18.92             & 0.48              & 0.428                & 17.57             & 0.34              & 0.646                & 21.98             & 0.64              & 0.392                & 23.85                & 0.68              & 0.523                & 20.58             & 0.54              & 0.497                \\
    {NVDiffrec}        & 18.21             & 0.52              & 0.381                & 18.85             & 0.45              & 0.429                & 23.38             & 0.87              & 0.170                & 25.29                & 0.84              & 0.283                & 21.43             & 0.67              & 0.316                \\
    {NVDiffrecMC+NeuS} & 15.81             & 0.42              & 0.531                & 19.20             & 0.45              & 0.454                & 27.97             & 0.84              & 0.165                & 24.73                & 0.86              & 0.255                & 21.93             & 0.64              & 0.351                \\
    {TensoIR}          & 20.27             & 0.59              & 0.346                & \bfseries 22.21   & 0.62              & 0.389                & 27.73             & 0.87              & 0.149                & \bfseries 26.10      & \bfseries 0.89    & 0.237                & 24.08             & 0.74              & 0.280                \\
    \bottomrule
    \end{tabular}
\end{adjustbox}
\caption{Novel-view synthesis results on multiple real and synthetic datasets.}
\label{tab:nvs}
\end{table*}

%% file: limitations.tex
\section{Limitations}
\label{sec:limitations}

We use a board with fiducial markers in our dataset to define a common coordinate frame for all images and environment maps.
The board is clearly visible in all environment maps and violates the assumption that the environment is a sphere with an infinite radius. %
Researchers should be able to bypass this limitation by including the board in the reconstruction.
Image acquisition and camera calibration processes are subject to some level of noise, which cannot be suppressed completely.

The presented baseline, Mitsuba+NeuS, optimizes geometry, and material and illumination sequentially. Also, it uses an analytic BRDF, which cannot represent all possible materials.

%% file: conclusion.tex
\section{Conclusion \& Future work}

We present the \emph{Objects With Lighting} dataset, a new resource designed for evaluating the reconstruction and relighting of 3D objects. This dataset enables precise assessment of inverse rendering methods in real-world scenarios by offering accurate, unclipped HDR images for lighting, and ground truth images for direct comparison with rendered images. Our dataset uniquely highlights existing issues with state-of-the-art methods, issues that synthetic datasets or novel-view synthesis tests may overlook.

The presented dataset, the code involved in the evaluation of all approaches, and the tools used to build the dataset are released to help the research community. We intend to continuously expand our dataset and refine the evaluation methods, enhancing their usefulness and value for the community.

%% file: appendix_dataset.tex
\newcommand{\contentlink}[1]{\hyperref[#1]{\ref{#1}\hspace{4mm}\nameref*{#1}\dotfill \pageref*{#1}}\\}

{
\bfseries
\noindent Contents

\vspace{1mm}
\noindent
\contentlink{sec:url-website}
\contentlink{sec:broader-impact}
\contentlink{sec:license}
\contentlink{sec:data-format}
\hspace*{5mm}\contentlink{sec:file-formats}
\hspace*{5mm}\contentlink{sec:dir-structure}
\hspace*{5mm}\contentlink{sec:coord-systems}
\contentlink{sec:data-collection}
\hspace*{5mm}\contentlink{sec:calibration}
\hspace*{5mm}\contentlink{sec:capture-protocol}
\hspace*{5mm}\contentlink{sec:post-processing}
\contentlink{sec:evaluation-metrics}
\contentlink{sec:relighting-experiments}
\hspace*{5mm}\contentlink{sec:methods}
\contentlink{sec:novel-view-synthesis-experiments}
\contentlink{sec:datasheet}
}

\section{URL and website}
\label{sec:url-website}
The dataset and code is available at \url{https://github.com/isl-org/objects-with-lighting}.
The code includes the scripts for evaluation and the tools that were used for building the dataset.

\section{Broader impact}
\label{sec:broader-impact}

The dataset is intended to evaluate the reconstruction and rendering of objects.
We think the risk of negative societal impact directly through our dataset is low.
The dataset does not contain personal information or offensive content.
However, the dataset is intended to help to improve photorealistic rendering of objects and there is a potential risk that this can be used to create artificial images that are not labelled as such.

\section{License}
\label{sec:license}
The dataset is licensed under the CDLA-permissive 2.0 license.
The source code is licensed under the Apache License 2.0.

\section{Data format}
\label{sec:data-format}

\subsection{File formats} 
\label{sec:file-formats}
Our dataset uses standard file formats for images and parameters to make it easy to use.
For images we use two file formats: the widely used PNG-format to store 24-bit RGB images for the input and ground truth images, and the RGBE or Radiance HDR format to store the equirectangular environment maps.
Both formats are lossless to avoid any artifacts when reading or writing the data.
The widely used OpenCV library can be used to read both formats.
Parameters are stored as text files and their format as well as example reading code using Numpy is provided in the repository.
For convenience we provide meshes generated with NeuS stored in the PLY format, which is supported by many 3D viewers and libraries.

\subsection{Directory structure} 
\label{sec:dir-structure}
The dataset is organized in a simple but extendable structure.
Each object is a directory with a unique name in the root of the dataset.
For each object we collect all data in a subfolder \texttt{test}, which stores the ground truth images for evaluation and a folder \texttt{inputs} with the data allowed to use for the reconstruction process.
The directory tree for an object with its core data files looks like this
\dirtree{%
.1 dataset.
.2 object name. 
.3 test. 
.4 gt\_image\_xxxx.png.
.4 gt\_camera\_xxxx.txt.
.4 gt\_env\_512\_rotated\_xxxx.hdr.
.4 gt\_exposure\_xxxx.txt.
.4 inputs. 
.5 image\_xxxx.png. 
.5 camera\_xxxx.png. 
.5 mask\_xxxx.png. 
.5 exposure.txt. 
.5 object\_bounding\_box.txt.. 
}

\subsection{Coordinate systems}
\label{sec:coord-systems}
Our dataset contains images, camera parameters, and environment maps, each with their respective coordinate system.
In the following, we describe these coordinate systems and how they are related.

\mypara{Cameras}
Cameras look in positive z-direction.
The intrinsic and extrinsic camera parameters can be used to directly project a 3D point $\XX$ to image space coordinates with 
\begin{equation}
    \xx = \KK \left( \RR \XX + \tt \right).
\end{equation}
$\xx$ is a homogeneous point describing a position in the image; $\KK$ is the upper triangular matrix with the camera intrinsics; and $\RR,\tt$ are the extrinsic parameters as rotation matrix and translation.

\mypara{Images}
The x-axis for images points to the right and the y-axis points down following the memory order. 
The coordinates $(x,y)$ of the top left corner are $(0,0)$.
The center of the first pixel is at $(0.5, 0.5)$.
The bottom right corner for an image with width $w$ and height $h$ is at $(w,h)$.

\begin{figure*}
    \centering
    \begin{subfigure}[t]{0.32\textwidth}
         \centering
         \adjustimage{width=\textwidth}{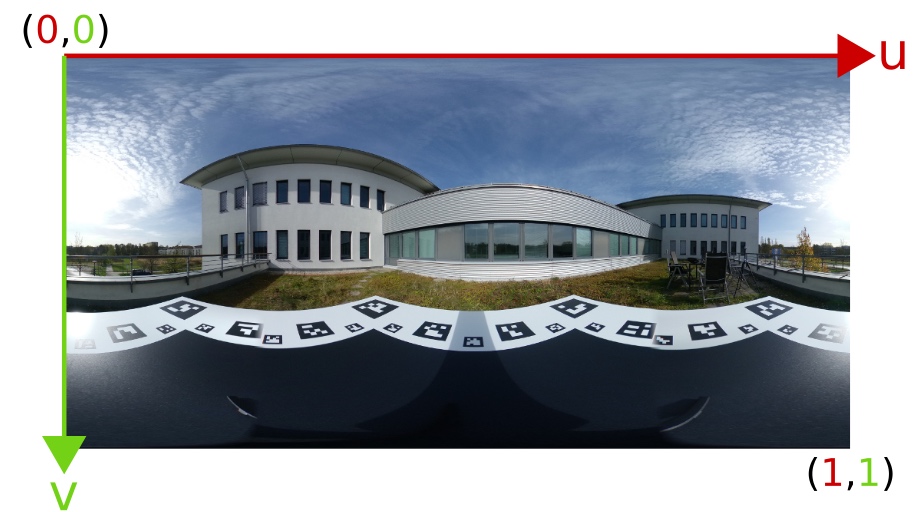}
         \caption{UV coordinates.}
         \label{fig:envmap-uv}
     \end{subfigure}
     \hfill
    \begin{subfigure}[t]{0.32\textwidth}
         \centering
         \adjustimage{width=\textwidth}{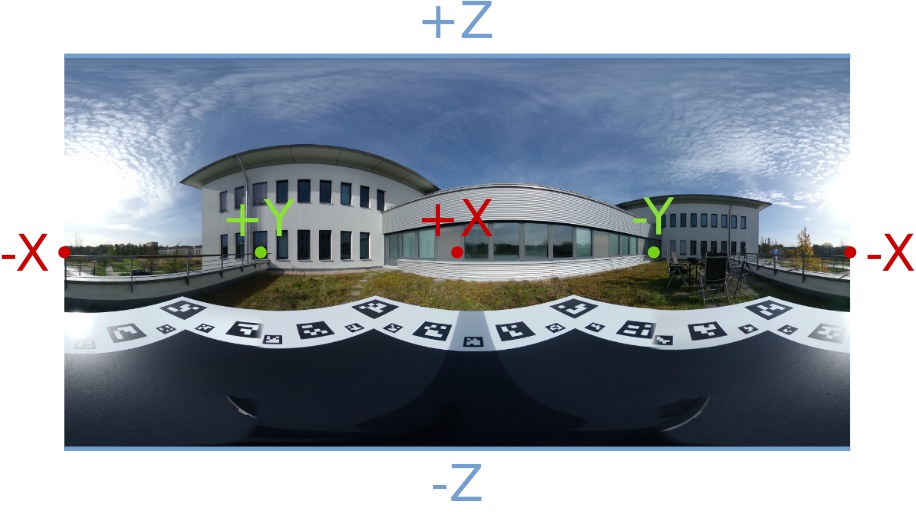}
         \caption{Spatial directions.}
         \label{fig:envmap-dirs}
     \end{subfigure}
     \hfill
    \begin{subfigure}[t]{0.3\textwidth}
         \centering
         \adjustimage{width=0.65\textwidth,keepaspectratio}{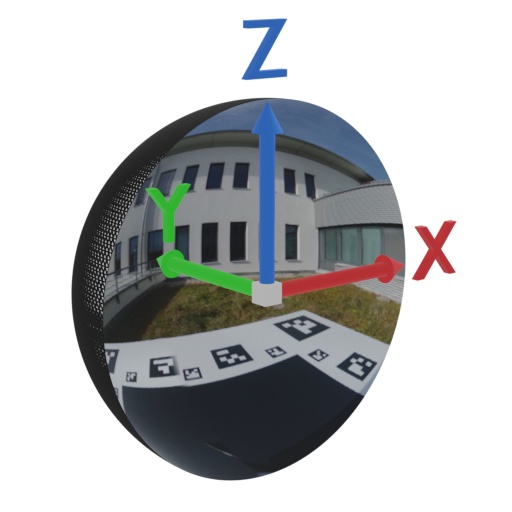}
         \caption{Left half of the environment map visualized as sphere.}
         \label{fig:envmap-sphere}
     \end{subfigure}
    \caption{Visualization of the coordinate systems used with environment maps.}
    \label{fig:envmap-coords}
\end{figure*}

\mypara{Environment maps}
Environment maps are stored as equirectangular images. We use a normalized coordinate system similar to regular images. 
The u-axis points to the right and the v-axis points down following the memory order. 
The coordinates $(u,v)$ of the top left corner are $(0,0)$.
The bottom right corner is at $(1,1)$ irrespective of the size of the environment map. 
This corresponds to the texture coordinate convention used by DirectX and is visualized in \fig{fig:envmap-uv}.

Directions map to the equirectangular image as shown in \fig{fig:envmap-dirs}. 
The direction +Z $(0,0,1)$ maps to the upper border of the environment map and -Z $(0,0,-1)$  to the lower border. 
+X $(1,0,0)$ maps to the center and -X $(-1,0,0)$ maps to the vertically centered point on the right and left border. 
+Y $(0,1,0)$ and -Y $(0,-1,0)$ map to the uv coordinates $(0.25,0.5)$ and $(0.75,0.5)$ respectively.
The environment map projected to a sphere with coordinate axes is visualized in \fig{fig:envmap-sphere}.

\section{Data collection}
\label{sec:data-collection}
\subsection{Calibration}
\label{sec:calibration}

Our dataset requires calibration of the linear color spaces of the cameras used, the intrinsic parameters of the cameras, and the geometric calibration of the extrinsics or camera poses.

\begin{figure*}
    \centering
    \begin{subfigure}[t]{0.30\textwidth}
         \centering
         \adjustimage{width=\textwidth}{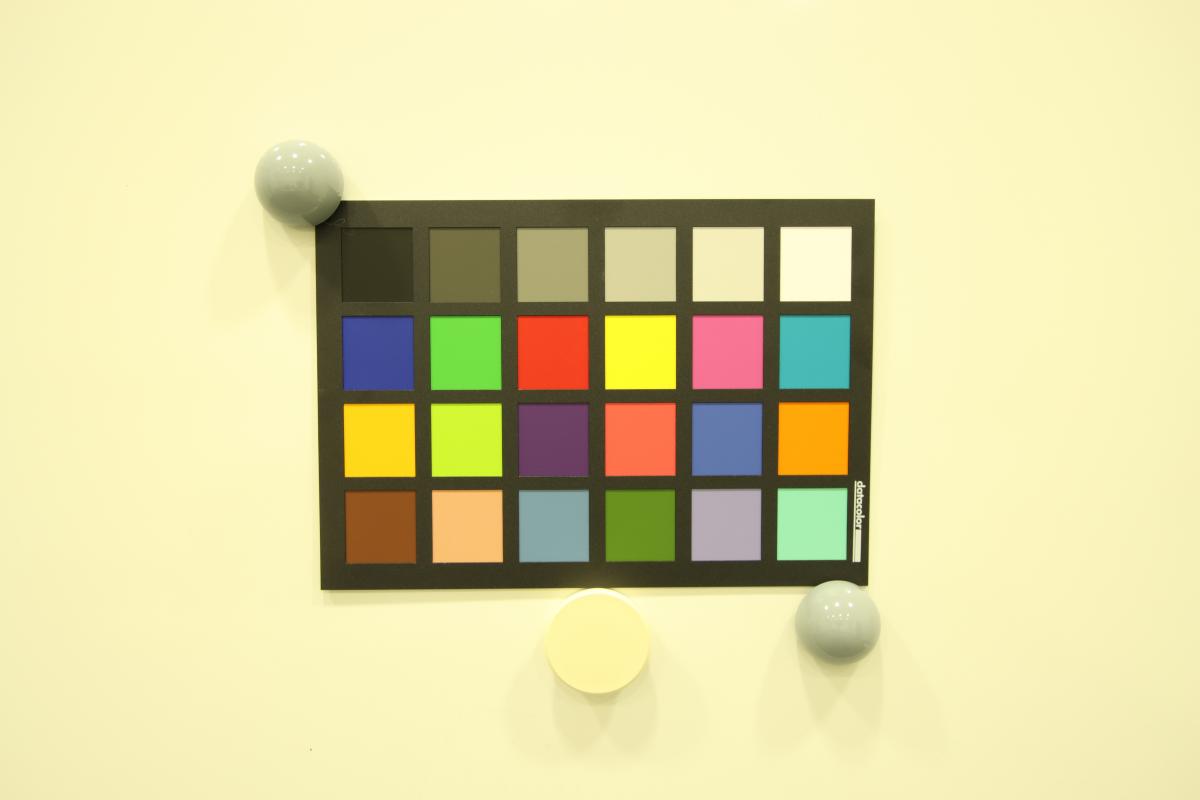}
         \caption{Color calibration target.}
         \label{fig:color-checker}
     \end{subfigure}
     \hfill
    \begin{subfigure}[t]{0.25\textwidth}
         \centering
         \adjustimage{width=0.83\textwidth}{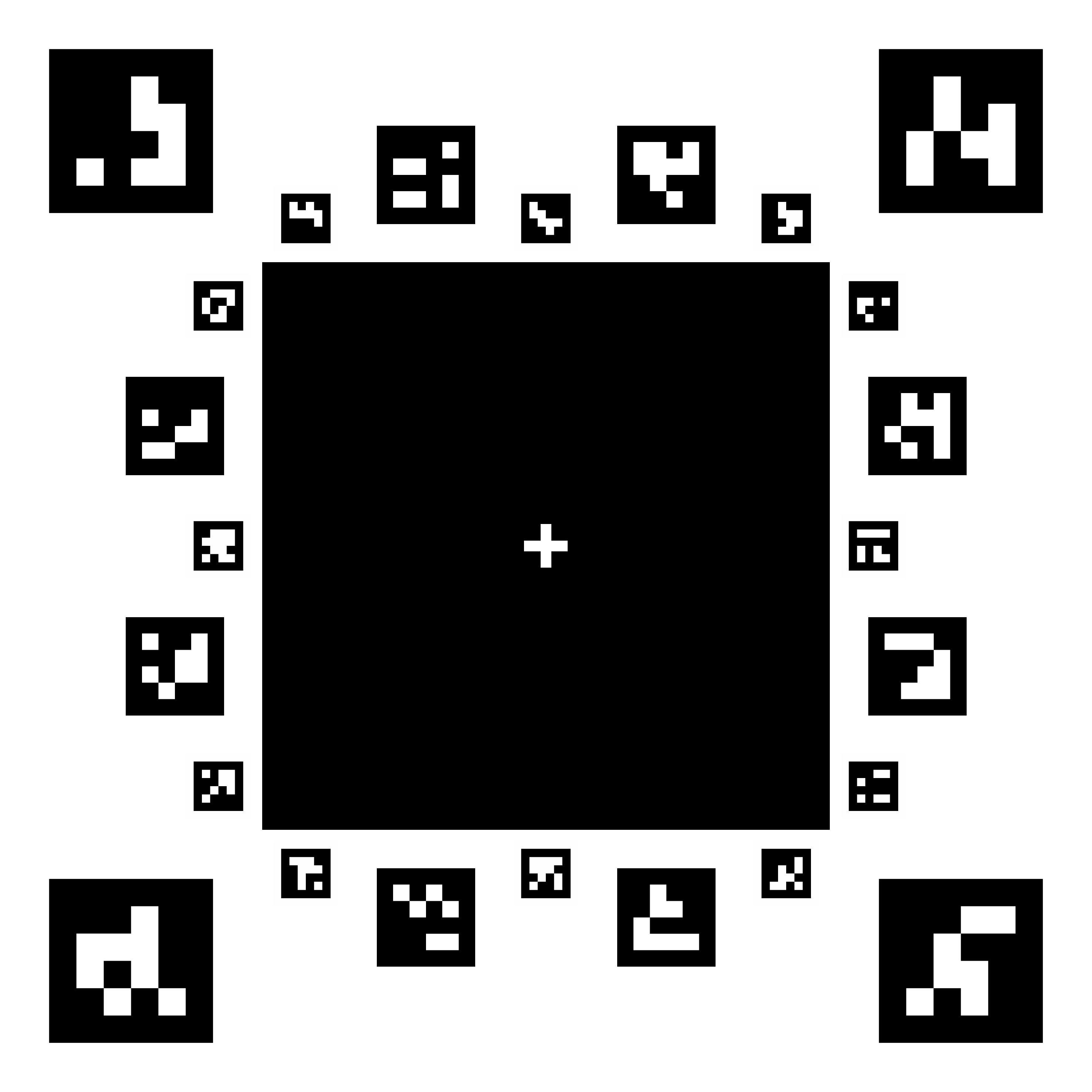}
         \caption{Board with AprilTags.}
         \label{fig:apriltag-board}
     \end{subfigure}
     \hfill
    \begin{subfigure}[t]{0.4\textwidth}
         \centering
         \adjustimage{width=\textwidth,keepaspectratio}{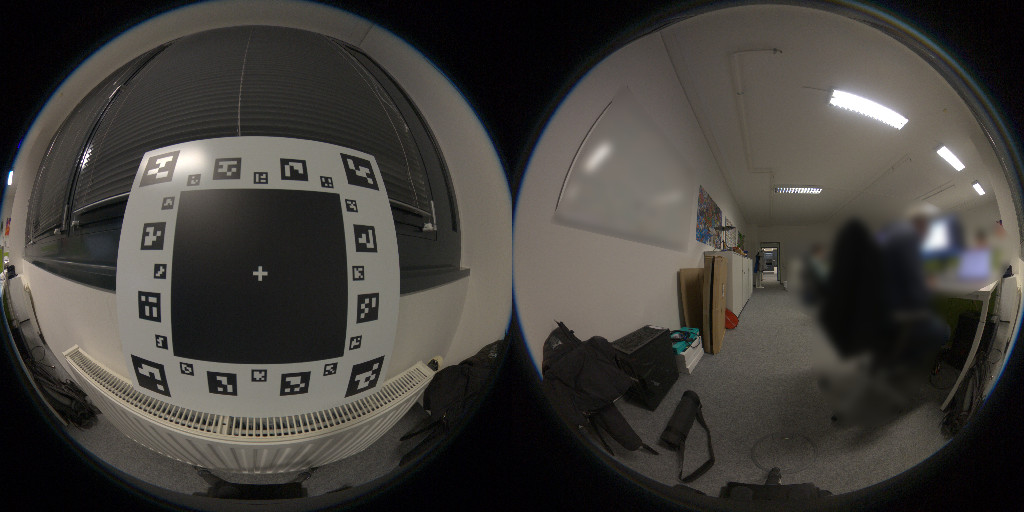}
         \caption{A calibration image for the Theta Z1.}
         \label{fig:theta-dfe-calib}
     \end{subfigure}
    \caption{Calibration targets for color and intrinsic camera parameters.}
    \label{fig:calibration-targets}
\end{figure*}

\mypara{Color}
Color calibration is important because we use two cameras that capture images in their individual color spaces.
We capture all images of the object with a Canon EOS90D but use a Ricoh Theta Z1 to capture environment maps.
To get all cameras into a common linear color space we use the color checker shown in \fig{fig:color-checker}.
We collect data for calibrating each camera by taking multiple images with different exposure values using the shutter speed. 
We keep the aperture fixed for both cameras (EOS 90D at $f/8.0$, Theta Z1 at $f/2.1$).
For ISO we always use 80 for the Theta Z1 but calibrate the EOS90D for multiple ISO values: 100, 500, 1000, 2000.
To compute the color transformation matrix we first merge the exposure brackets to HDR images and compute the least squares $3\times3$ color transform that maps the HDRIs from the EOS90D to the linear color space of the Theta Z1.

\mypara{Camera Intrinsics}
To estimate the camera intrinsics we use the calibration pipeline from OpenCV.
We use the OpenCV pinhole camera model but limit the radial distortion to a single parameter.
For calibrating the fisheye cameras of the Theta Z1 we use the camera model presented in \cite{kannala2006generic}.
We use a board with AprilTag markers as shown in for the calibration.
One of the images used for calibration is shown in \fig{fig:theta-dfe-calib}.

\begin{figure*}
    \centering
    \adjustimage{width=\textwidth}{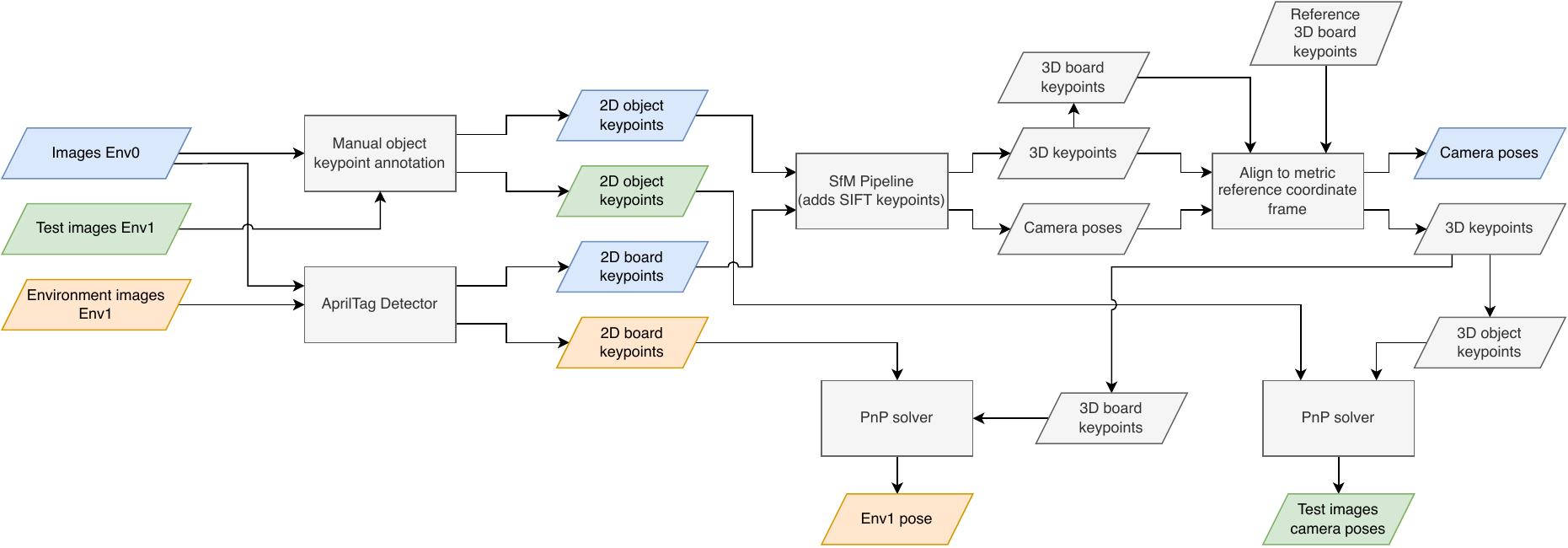}
    \caption{Overview of the camera pose calibration.%
     We show the process for aligning test images and the environment map for an environment \emph{Env1} along with the alignment of images of an \emph{Env0} which are used as input for the tested methods.
     Colors indicate the association of intermediate data and outputs to the input images where disambiguation is needed.}
    \label{fig:campose-workflow}
\end{figure*}

\begin{figure*}
    \centering
    \begin{subfigure}[t]{0.60\textwidth}
         \centering
         \adjustimage{width=\textwidth}{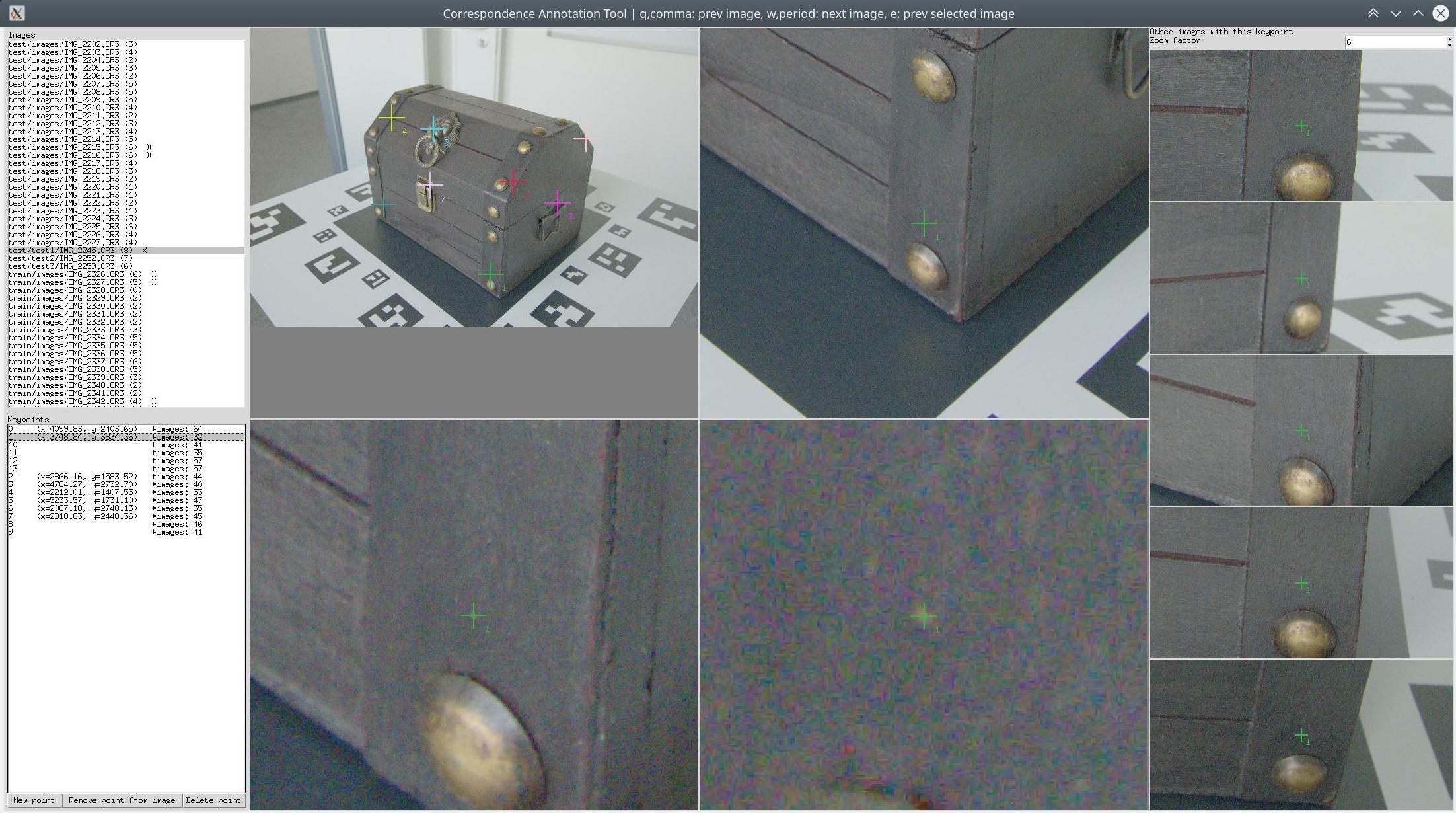}
         \caption{Tool for annotating object keypoints.}
         \label{fig:annotationtool}
     \end{subfigure}
     \hfill
    \begin{subfigure}[t]{0.35\textwidth}
         \centering
         \adjustimage{width=\textwidth,keepaspectratio}{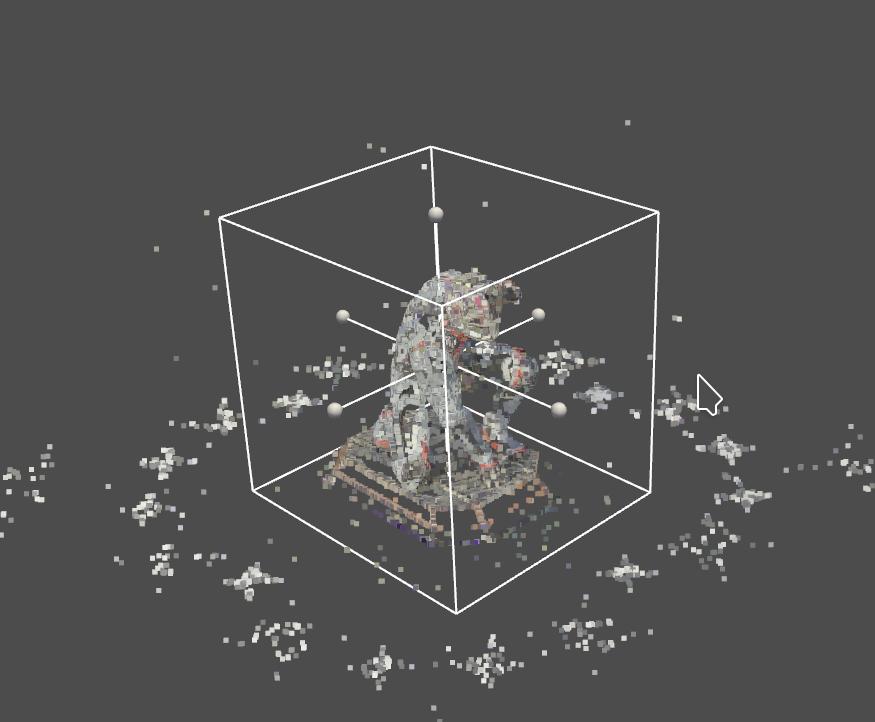}
         \caption{Bounding box annotation.}
         \label{fig:bbox}
     \end{subfigure}
    \caption{Annotation tools to define keypoints on the object and the bounding box.}
    \label{fig:bbox-annotation}
\end{figure*}

\mypara{Camera poses}
Calibration of the camera poses is crucial since we want to compare images of the same object taken in different environments and also relight them with the measured illumination in the form of environment maps.
See \fig{fig:campose-workflow} for an overview of the calibration process.
We use COLMAP \cite{colmap} with a mixture of automatically detected SIFT keypoints, keypoints from a board with AprilTag markers as seen in \fig{fig:apriltag-board}, and expert annotated keypoints.
For each scene, we detect the AprilTags in all images and inject the corners into the Structure from Motion pipeline as constant points.
Knowing the geometry of the AprilTag board this allows us to define a metric coordinate system with the center of the board as the origin.
To align the environment maps to this coordinate system we detect the AprilTags in the fisheye images by temporarily converting them into 8 pinhole projections for the detector.
With the detected corner points of the markers, we then solve a PnP problem to compute the pose for the omnidirectional camera.
In the last step, we have to align the test images taken in a different environment to the object in the environment which we use for the reconstruction.
Since the only common structure is the object itself we need to find correspondences on the object.
We found that automatic detections are not reliable, adding markers on the objects is in many cases not feasible, and gluing the object to the board does not allow for easy transport. 
Thus, we resort to manual annotations that relate the images of the same objects in different environments. \fig{fig:annotationtool} shows our correspondence annotation tool specialized for this task.
We annotate $\geq 10$ points for each object in the test images and the images used for reconstruction.
We distribute the annotated points over the object and use these points in the Structure from Motion pipeline too.
To register the test images we solve a PnP problem to compute the camera poses of the test images.

\subsection{Capture protocol}
\label{sec:capture-protocol}

We describe the capture process for a scene in our dataset.

\mypara{Step 1} We start with capturing a reference environment map. 
We place the 360 camera in the center of the marker board and capture exposure brackets with 11 images at different shutter speeds.
In outdoor environments, we shoot additional images with a custom ND filter to get the full unclipped range of the sunlight.

\mypara{Step 2} After capturing a reference for the environment, we remove the 360 camera and place the object in the center of the board.
We take 42 to 67 photos of each object setting shutter speed and ISO such that we avoid overexposure.
We use the same settings for all photos of the object meant as input for the reconstruction.

\mypara{Step 3} We shoot the ground truth images.
To be able to change the exposure values afterward and to enable evaluation of linear images in the future, we use a tripod and capture exposure brackets with 7 images, which we merge into an HDR image.
We repeat the process three times to create ground truth images from three different angles.
Each time we mark the position of the tripod for Step 4.

\begin{figure*}
    \centering
    \adjustimage{width=0.32\textwidth}{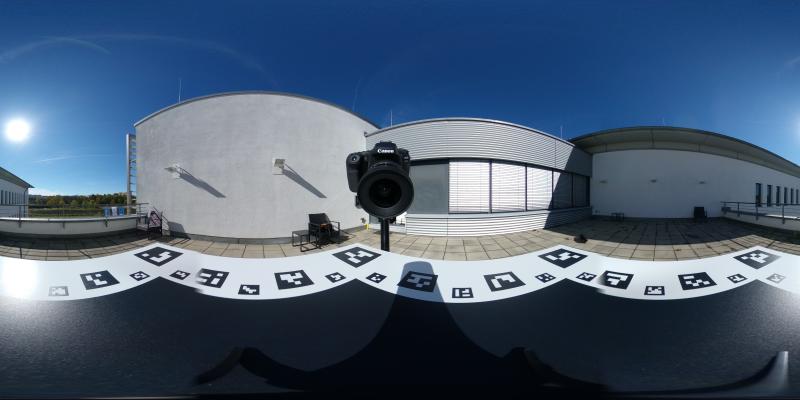}
    \adjustimage{width=0.32\textwidth}{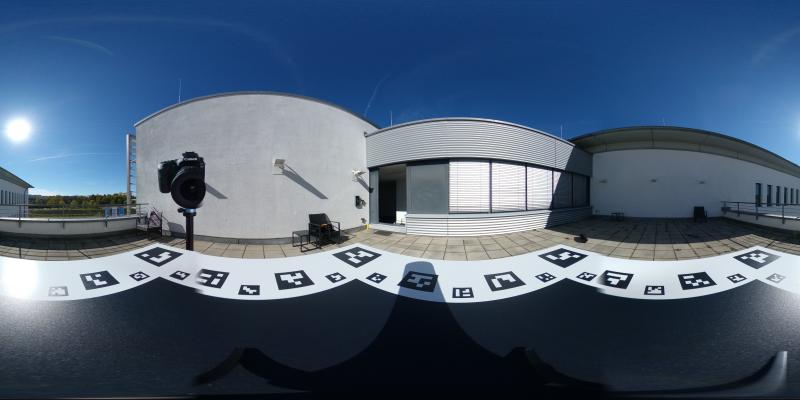}
    \adjustimage{width=0.32\textwidth}{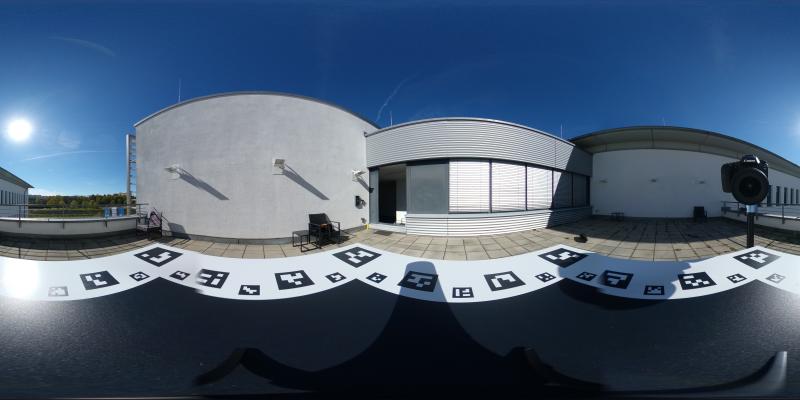}
    \caption{The camera and tripod are part of the environment map for test images because they affect the illumination.}
    \label{fig:testenvs}
\end{figure*}

\mypara{Step 4}
We replace the object with the 360 camera and capture three HDR environment maps as described in Step 1.
For each environment map we put the tripod with the DSLR camera on one of the previously marked positions.
This allows us to capture the environment as in Step 2 in which the tripod and camera can act as occluders and affect the illumination.
An example of the environment maps captured in this step is shown in \fig{fig:testenvs}.

\subsection{Post processing}
\label{sec:post-processing}

Besides the calibration, we apply post-processing steps to make the data easier to use and reduce errors.
For environment maps, we stitch the dual fisheye images produced by the 360 camera and provide directly equirectangular images that use the coordinate systems described in \sectn{sec:coord-systems} and are aligned with the pinhole cameras of the input and ground truth images.

Many methods require the bounds of the object to be reconstructed to work efficiently.
We manually annotate 3D bounding boxes for all objects using the sparse reconstruction obtained from COLMAP as seen in \fig{fig:bbox-annotation}.

Additionally, we provide approximate foreground masks for the input images.
To this end, we reconstruct the objects with NeuS and manually clean the mesh.
The masks are generated by projecting the mesh to each input image. 
For the foreground masks used in the evaluation with the test images, we use expert annotations instead.

The last post-processing step is the downsampling of the generated images.
For the input and test images, we reduce the resolution from $6960\times 4640$ to $1746\times 1165$ to facilitate working with the data and to alleviate any errors.
For the environment maps, we downsample from $6720\times 3360$ to $1024\times 512$ for the same reasons.

\section{Evaluation metrics}
\label{sec:evaluation-metrics}

We use the standard metrics peak signal-to-noise ratio (PSNR), structural similarity index measure (SSIM), and learned perceptual image patch similarity (LPIPS) \cite{zhang2018perceptual} to compare rendered images with their corresponding ground truth.
We briefly describe how we use these metrics in our evaluation script.

For PSNR and SSIM we use the implementation in scikit-image \cite{scikitimage}.
We convert and scale our 8-bit tone-mapped RGB images to floating point images with range $[0..1]$ before computing the two metrics.
In addition, we set background pixels to 0 using the ground truth foreground mask.
We compute the PSNR pixel-wise for all channels and aggregate values over the valid pixels defined by the foreground mask.
For SSIM we use a sliding window with a size of $7\times 7$ pixels and compute the similarity for each pixel and channel and aggregate over all valid pixels.
The values we report are the average PSNR and SSIM over the valid pixels.

For LPIPS we use the implementation provided by the authors.
We transform our images to floating point images with range $[-1..1]$ and set background pixels to 0. 
To account only for the valid pixels we compute the differences per pixel by scaling the feature difference maps to the original image size.
This corresponds to passing the option \texttt{spatial=True}.
We chose AlexNet as the network.
The reported value is the average of the per-pixel loss image for the valid pixels.

%% file: appendix_experiments.tex
\section{Relighting experiments}
\label{sec:relighting-experiments}
\begin{figure*}[p]
    \centering
    \adjustimage{width=0.97\textwidth}{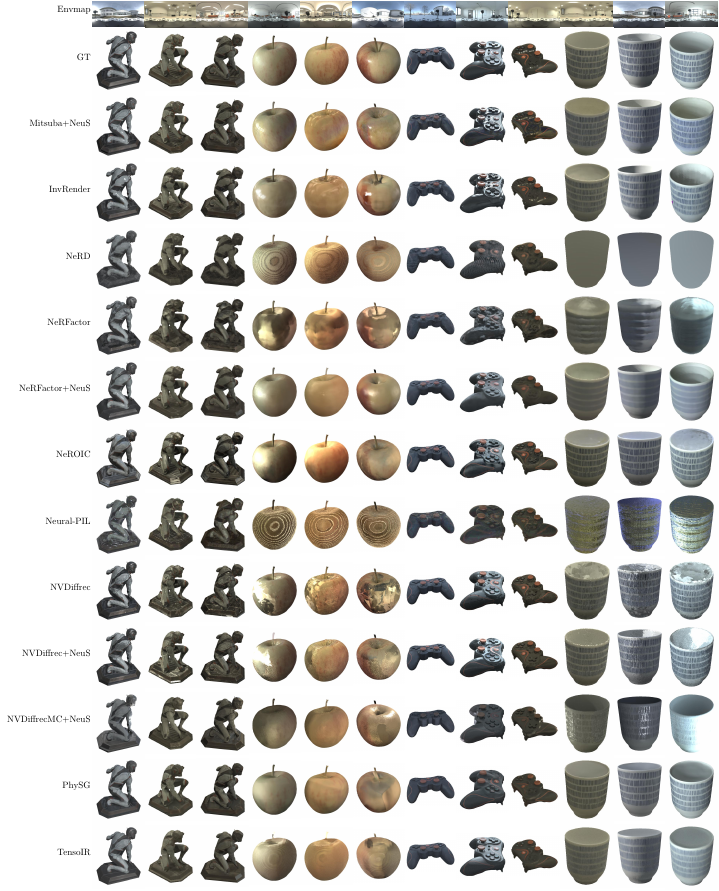}
    \caption{Qualitative examples for relighting on our dataset.
    The first row shows the environment map used for relighting.
    For each object, we show one of the test images of the 3 lighting categories.
    The first image for each object corresponds to the reconstruction environment. Each object has been captured in highly different lighting conditions: outdoor, indoor, and indoor with artificial light.
    }
    \label{fig:ord_examplesA}
\end{figure*}

\begin{figure*}[p]
    \centering
    \adjustimage{width=0.85\textwidth}{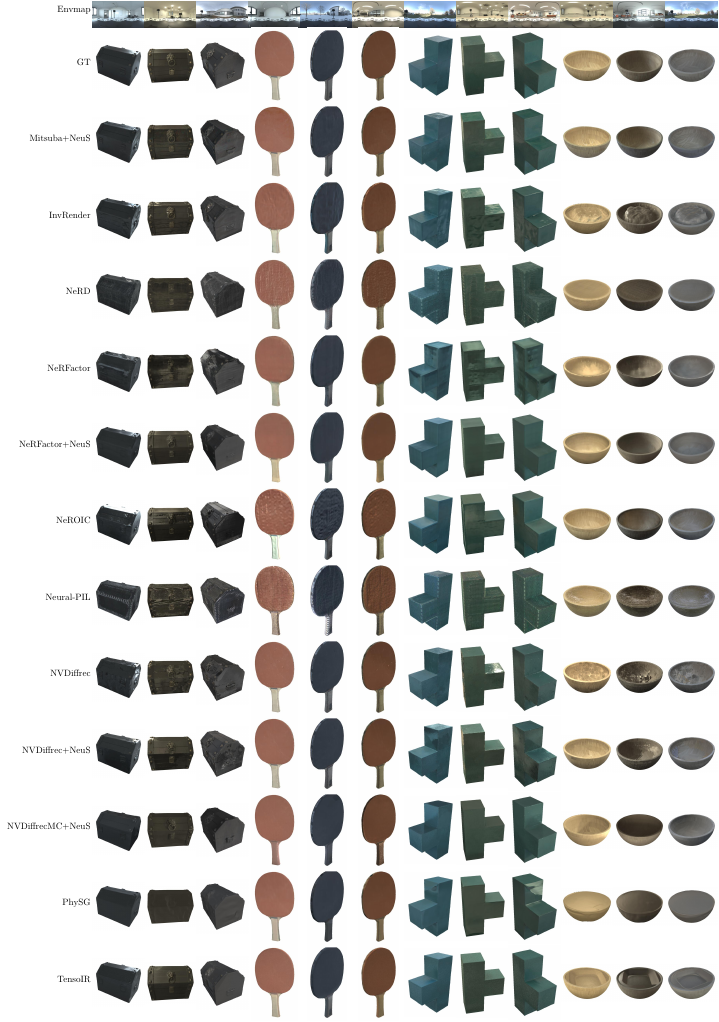}
    \caption{Qualitative examples for relighting on our dataset.
    The first row shows the environment map used for relighting.
    For each object, we show one of the test images of the 3 lighting categories.
    The first image for each object corresponds to the reconstruction environment. 
    Each object has been captured in highly different lighting conditions: outdoor, indoor, and indoor with artificial light. 
    }
    \label{fig:ord_examplesB}
\end{figure*}

\begin{figure*}[p]
    \centering
    \adjustimage{width=0.9\textwidth}{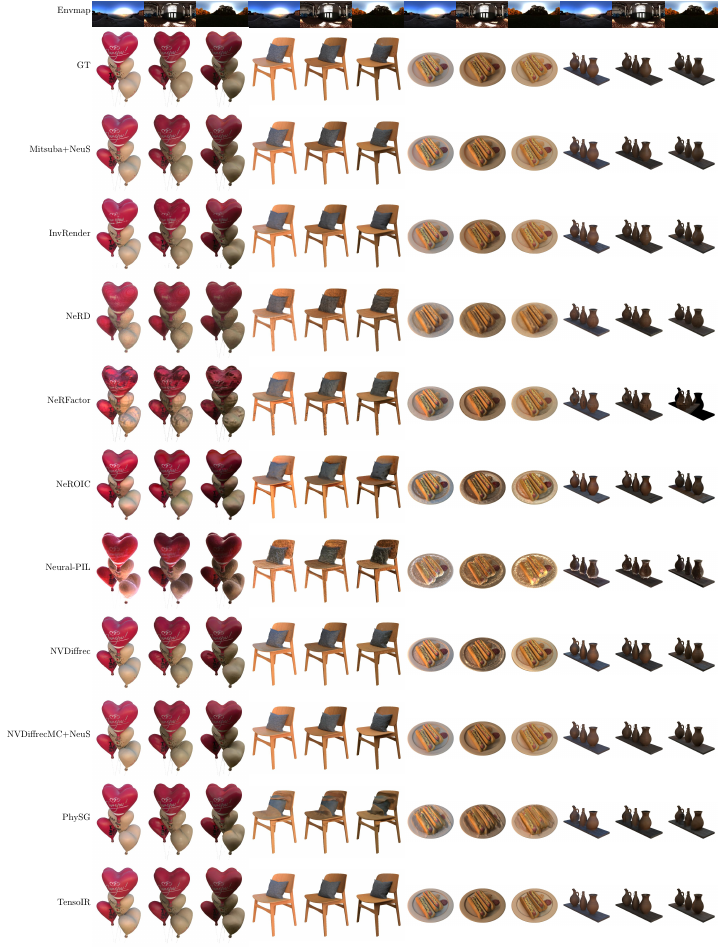}
    \caption{Qualitative examples for relighting on the Synthetic4Relight dataset.
    The first row shows the environment map used for relighting.
    For each object, we show one of the test images.
    The first image for each object corresponds to the reconstruction environment.
    }
    \label{fig:synth4relight_examples}
\end{figure*}

We provide additional details for the relighting experiments on our dataset and Synthetic4Relight.
Quantitative results on our dataset are reported in \tab{tab:object-relighting} and \tab{tab:synth4relight}.
We show qualitative examples for each object of the Objects With Lighting dataset in \fig{fig:ord_examplesA} and \fig{fig:ord_examplesB}.
For Synthetic4Relight we show examples in \fig{fig:synth4relight_examples}.

\begin{figure*}
   \centering
    \begin{subfigure}[t]{0.40\textwidth}
        \centering
        \adjustimage{width=0.4\textwidth}{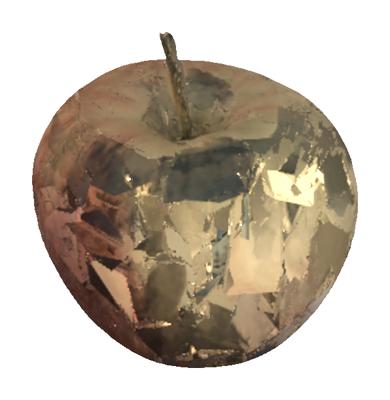}
        \adjustimage{width=0.4\textwidth}{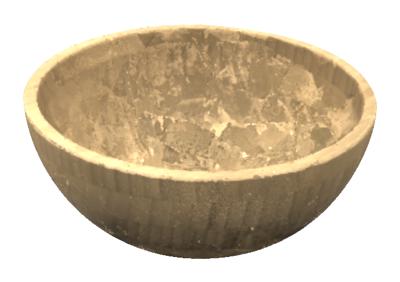}
        \caption{NVDiffrec}
    \end{subfigure}
    \hfill
    \begin{subfigure}[t]{0.16\textwidth}
        \centering
        \adjustimage{width=\textwidth}{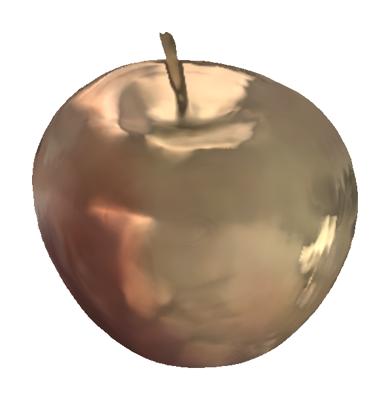}
        \caption{NeRFactor}
    \end{subfigure}
    \hfill
    \begin{subfigure}[t]{0.14\textwidth}
        \centering
        \adjustimage{width=\textwidth}{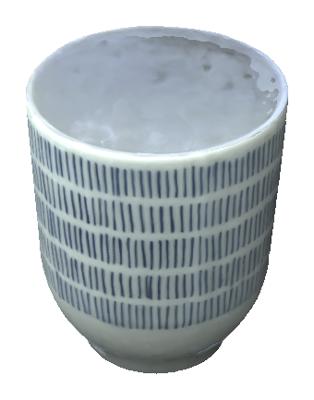}
        \caption{NeROIC}
    \end{subfigure}
    \hfill
    \begin{subfigure}[t]{0.16\textwidth}
        \centering
        \adjustimage{width=\textwidth}{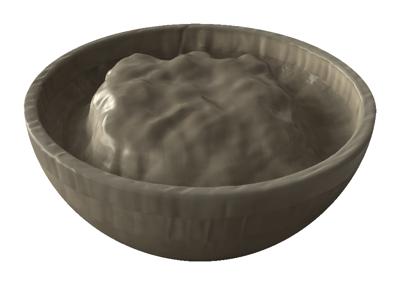}
        \caption{InvRender}
    \end{subfigure}
    \hfill
    \caption{Artifacts related to failures in shape reconstruction. \textbf{(a)} and \textbf{(b)}: noisy shapes lead to problems in the relighted images. \textbf{(c)} and \textbf{(d)}: concave shapes are problematic for most methods.}
    \label{fig:shape-problems}
\end{figure*}

We can identify multiple failure modes on our dataset.
The most obvious artifacts stem from the shape reconstruction.
These are especially visible for NVDiffrec which uses an explicit surface representation and produces noisy surfaces.
Another problem is related to the reconstruction of concave shapes, which can be observed for multiple methods.
We show examples of these problems in \fig{fig:shape-problems}.

\begin{figure*}
   \centering
    \begin{subfigure}[t]{0.24\textwidth}
        \centering
        \adjustimage{width=\textwidth}{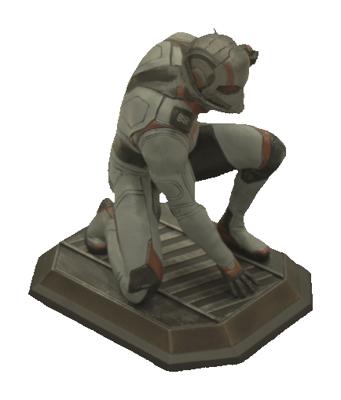}
        \adjustimage{width=\textwidth}{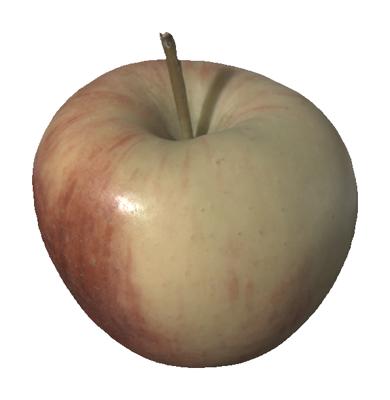}
        \caption{GT}
    \end{subfigure}
    \hfill
    \begin{subfigure}[t]{0.24\textwidth}
        \centering
        \adjustimage{width=\textwidth}{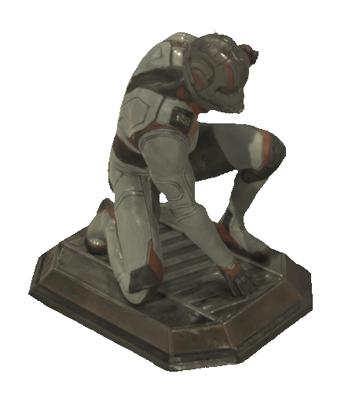}
        \adjustimage{width=\textwidth}{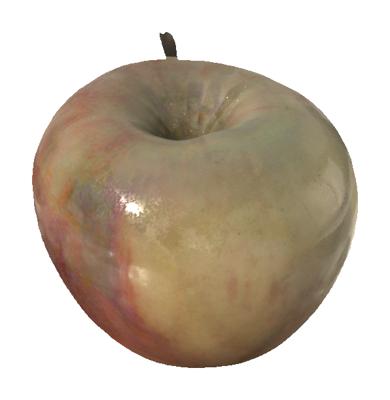}
        \caption{Mitsuba+NeuS}
    \end{subfigure}
    \hfill
    \begin{subfigure}[t]{0.24\textwidth}
        \centering
        \adjustimage{width=\textwidth}{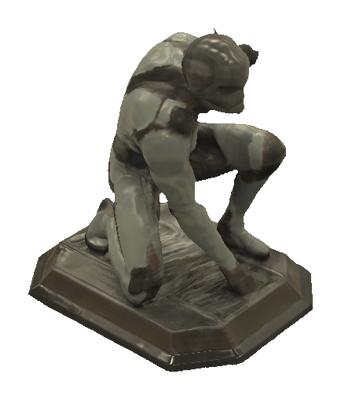}
        \adjustimage{width=\textwidth}{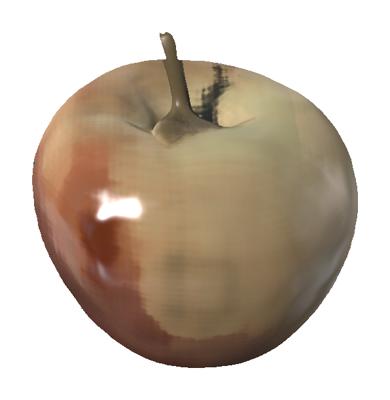}
        \caption{InvRender}
    \end{subfigure}
    \hfill
    \begin{subfigure}[t]{0.24\textwidth}
        \centering
        \adjustimage{width=\textwidth}{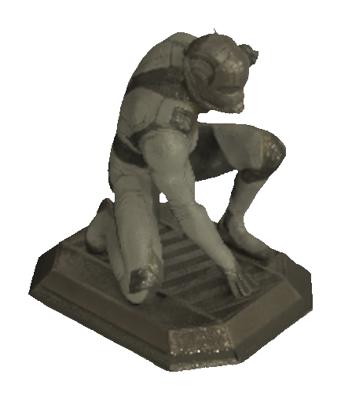}
        \adjustimage{width=\textwidth}{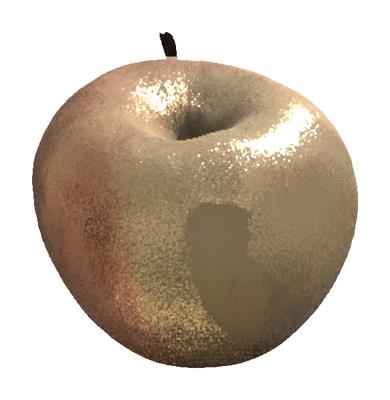}
        \caption{NVDiffrecMC+NeuS}
    \end{subfigure}
    \caption{Examples where objects appear too glossy compared to the ground truth.}
    \label{fig:glossiness-problems}
\end{figure*}

Another failure mode can be observed with respect to the estimated materials.
Methods like our baseline, InvRender, and NVDiffrecMC tend towards overestimating the glossiness of objects, which can be observed by comparing highlights in the ground truth and rendered images.
We show examples in \fig{fig:glossiness-problems}.

\begin{figure*}
   \centering
    \begin{subfigure}[t]{0.24\textwidth}
        \centering
        \adjustimage{width=\textwidth}{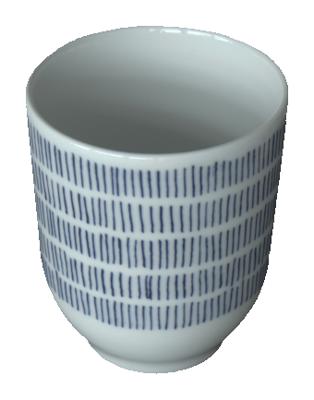}
        \caption{GT}
    \end{subfigure}
    \hfill
    \begin{subfigure}[t]{0.24\textwidth}
        \centering
        \adjustimage{width=\textwidth}{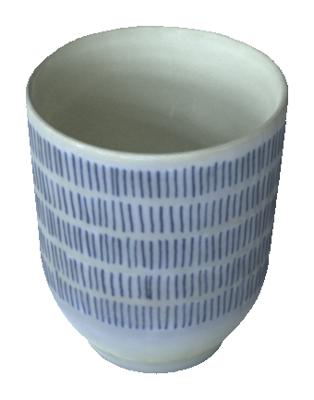}
        \caption{Mitsuba+NeuS}
    \end{subfigure}
    \hfill
    \begin{subfigure}[t]{0.24\textwidth}
        \centering
        \adjustimage{width=\textwidth}{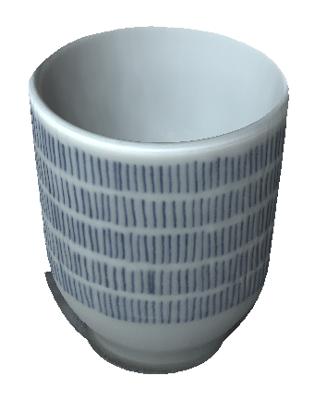}
        \caption{NVDiffrec+NeuS}
    \end{subfigure}
    \caption{Methods that do not account for the visibility of the illumination have a disadvantage in reproducing shadows and are also prone to bake shadows into the material.}
    \label{fig:shadow-problems}
\end{figure*}

Besides the reconstruction, the employed rendering technique is crucial to produce realistic images.
This can be seen on the porcelain mug in \fig{fig:shadow-problems} for which NVDiffrec misses to recreate the shadow. Methods like Mitsuba which use a more sophisticated rendering process account for the visibility of the illumination and can more faithfully recover generate the shadow.

\subsection{Methods}
\label{sec:methods}

\mypara{Mitsuba+NeuS}
We describe our baseline method in \sectn{sec:method} and provide the source code in the accompanying GitHub repository of the dataset to reproduce results.
We use for all experiments the same parameters.
Reconstruction of one scene of our dataset takes about 10 hours on an Nvidia A6000.
The majority of the time is spent on the surface reconstruction using NeuS~\cite{wangNeuSLearningNeural2021}.
The material reconstruction using Mitsuba~\cite{Mitsuba3} takes only about 15 minutes.

\mypara{InvRender}
The method is described in \cite{zhang2022invrender}.
We provide instructions and scripts for reproducing results in our code release. 
All experiments use the same parameters. For each scene, the geometry reconstruction stage takes about 12 hours on an Nvidia A100. The illumination and visibility stage takes about 2 hours. The material optimization stage takes about 3 hours.

\mypara{NeRD}
The method is described in \cite{bossNeRDNeuralReflectance2021}.
We provide instructions and scripts for reproducing results in our code release. We use the same parameters for all scenes in the Objects With Lighting dataset. For DTU and BMVS, we scale the scene to fit within a unit sphere. For each scene, it takes about 20 hours on an Nvidia RTX 3090. For relighting, we've modified the code to use our tone mapping equation, and to align the environment map to our coordinate system. We save relighting output as hdr files to apply exposure correction and color balancing before evaluation.

\mypara{NeRFactor(+NeuS)}
The method is described in \cite{zhangNeRFactorNeuralFactorization2021}.
We provide instructions and scripts for reproducing results in our code release.
The coarse NeRF reconstruction stage takes about 16-24 hours on an Nvidia A6000 and depends on the number of images and resolution of the scene.
Geometry extraction takes about 20 minutes for each image, which can be parallelized. Further joint optimization takes about 40 minutes. We keep parameters constant for all experiments but vary the image resolution to improve reconstruction results. If combined with NeuS we can skip the NeRF reconstruction.

\mypara{NeROIC}
The method is described in \cite{neroic}.
We provide instructions and scripts for reproducing results in our code release. We use the same parameters for all scenes in the Objects With Lighting dataset. For DTU and BMVS, we scale the scene to fit within a unit sphere. For each scene, it takes about 10 hours to run geometry reconstruction, 1 hour to extract normals, and about 5 hours to optimize the rendering network on an Nvidia RTX 3090.

\mypara{Neural-PIL}
The method is described in \cite{bossNeuralPILNeuralPreIntegrated2021}.
We provide instructions and scripts for reproducing results in our code release. We use the same parameters for all scenes in the Objects With Lighting dataset. For DTU and BMVS, we scale the scene to fit within a unit sphere. For each scene, it takes about 24 hours to train on an Nvidia RTX 3090.  For relighting, we've modified the code to use our tonemapping equation, and to align the environment map to our coordinate system. We save relighting output as hdr files to apply exposure correction and color balancing before evaluation.

\mypara{NVDiffrec(+NeuS)}
The method is described in \cite{munkbergExtractingTriangular3D2021}.
We provide instructions and scripts for reproducing results in our code release. We use the same parameters for all the scenes. For each scene, it takes about 1 hour for the geometry reconstruction pass, and another 1 hour to optimize the materials on an Nvidia RTX 3090. If combined with NeuS, we can skip the geometry reconstruction pass.

\mypara{NVDiffrecMC+NeuS}
The method is described in \cite{hasselgren2022nvdiffrecmc}.
We provide instructions and scripts for reproducing results in our code release. 
When training from scratch, each scene takes about 4 hours to train a coarse mesh and another 4 hours to optimize materials on an Nvidia A100. If training with the meshes generated with NeuS, we first use Blender to do UV-unwrap for each mesh, and optimize the material with this fixed mesh. Each experiment only needs about 4 hours in this configuration. All experiments use the same parameters. 

\mypara{PhySG}
The method is described in \cite{zhangPhySGInverseRendering2021}.
We provide instructions and scripts for reproducing results in our code release. We modify PhySG's environment map convention to be consistent with that of our data. We keep the default hyper-parameters of PhySG and train from scratch. Each scene takes about 12 hours to train on an Nvidia RTX 2080Ti. For relighting, we first overfit a Spherical Gaussian (SG) representation (128 SG using the default PhySG setting) for each environment lighting, and then use the SG-based lighting to relight the reconstructed 3D model.

\mypara{TensoIR}
The method is described in \cite{Jin2023TensoIR}.
We provide instructions and scripts for reproducing results in our code release.
We use the default parameters set by the author. We use masked input images with white backgrounds for training, similar to TensoIR's own synthetic dataset. For DTU and BMVS, we scale the scene to fit within a unit sphere before training. For each scene, the training takes about 15 hours on a single Nvidia RTX3090. For novel view synthesis evaluation, we report the results from the volume rendering network, which has higher PSNRs than the BRDF rendering network.

\section{Novel view synthesis experiments}
\label{sec:novel-view-synthesis-experiments}

\begin{figure*}[p]
    \centering
    \adjustimage{width=0.9\textwidth}{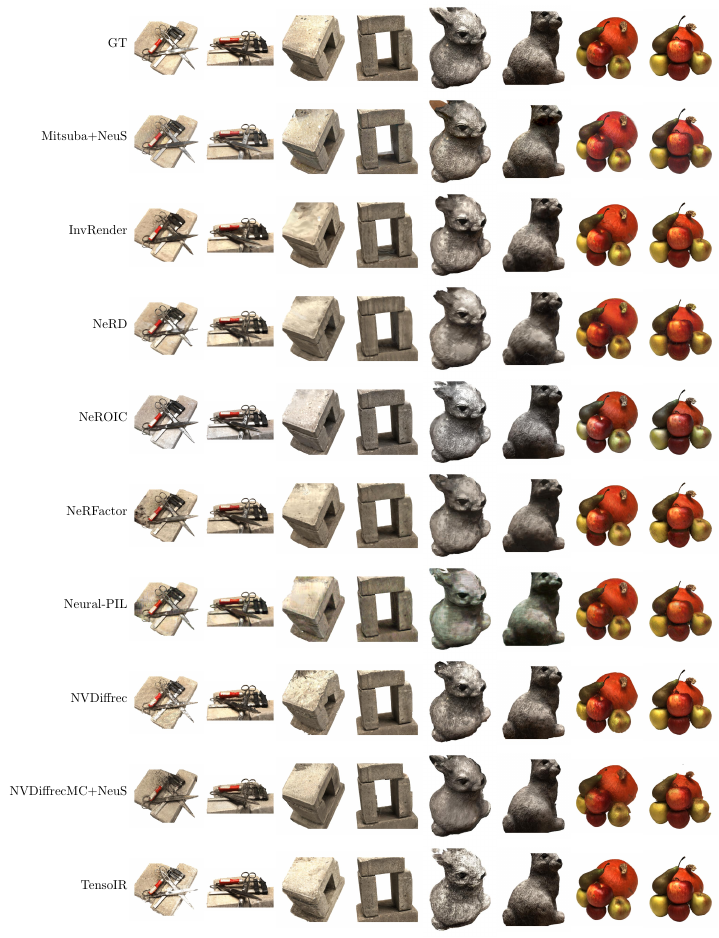}
    \caption{Qualitative examples for novel view synthesis on scenes from the DTU dataset.
    For each scene, we show two novel views.
    }
    \label{fig:dtu_examples1}
\end{figure*}

\begin{figure*}[p]
    \centering
    \adjustimage{width=0.9\textwidth}{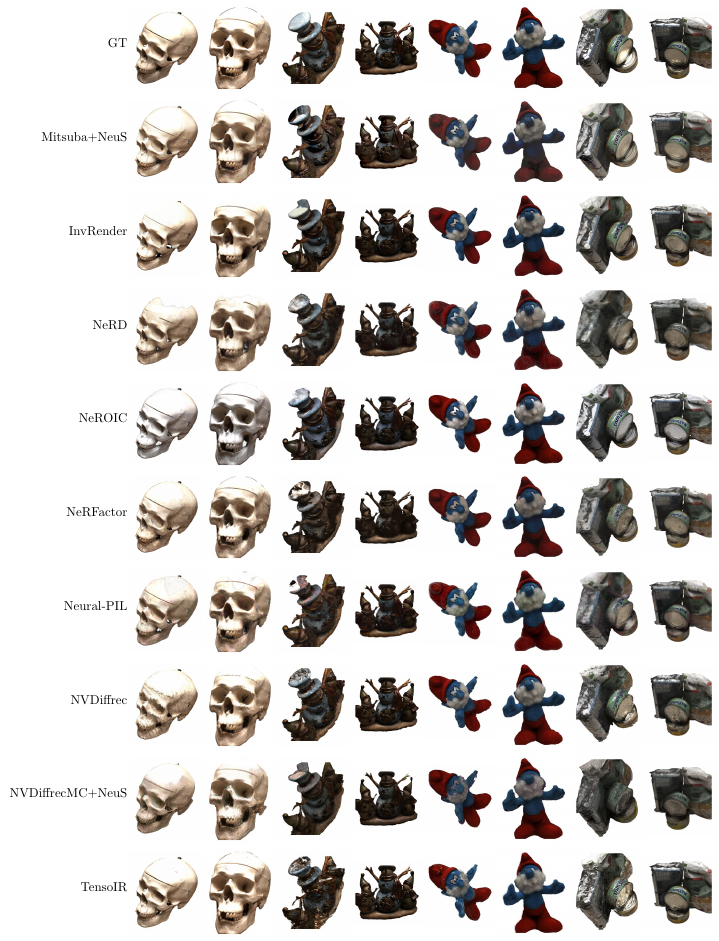}
    \caption{Qualitative examples for novel view synthesis on scenes from the DTU dataset.
    For each scene, we show two novel views.
    }
    \label{fig:dtu_examples2}
\end{figure*}

\begin{figure*}[p]
    \centering
    \adjustimage{width=\textwidth}{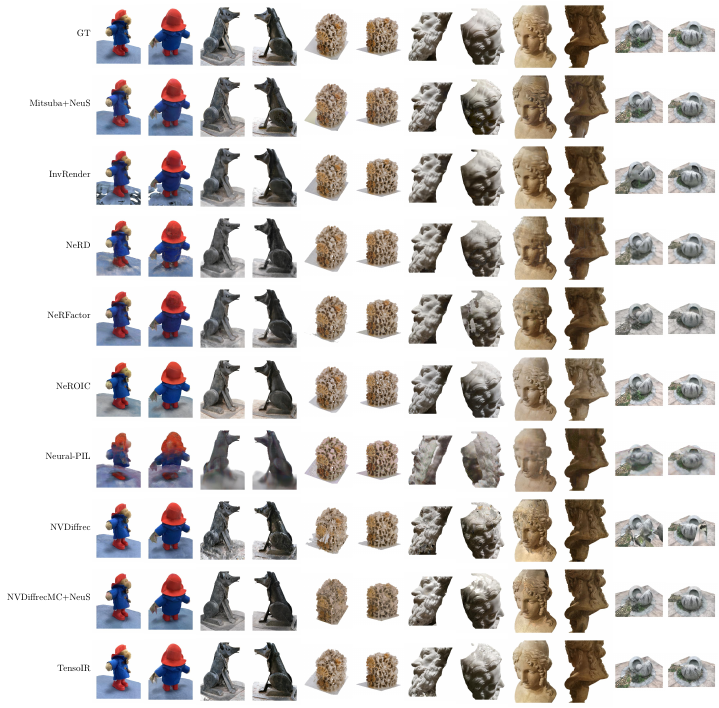}
    \caption{Qualitative examples for novel view synthesis on scenes from the BlendedMVS dataset.
    For each scene, we show two novel views.
    }
    \label{fig:bmvs_examples}
\end{figure*}

\begin{figure*}[p]
    \centering
    \adjustimage{width=\textwidth}{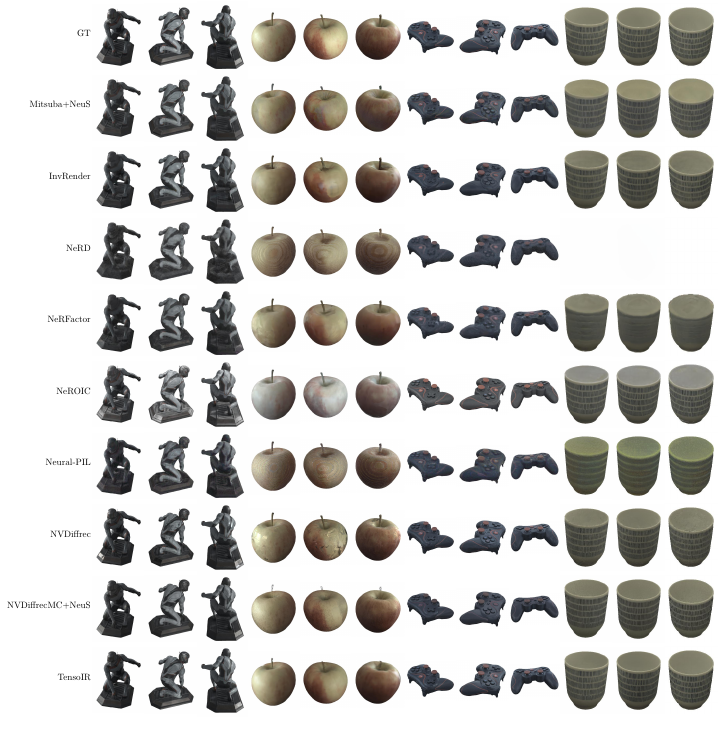}
    \caption{Qualitative examples for novel view synthesis on our dataset for the first environment.
    For each scene, we show three novel views. Note that NeRD failed to reconstruct the porcelain mug scene.
    }
    \label{fig:ord_nvs_examplesA}
\end{figure*}

\begin{figure*}[p]
    \centering
    \adjustimage{width=\textwidth}{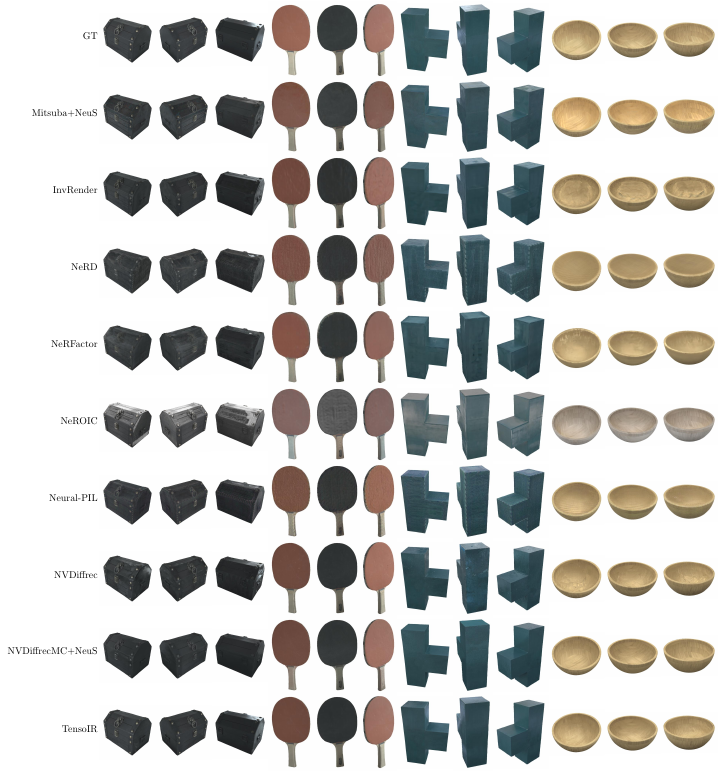}
    \caption{Qualitative examples for novel view synthesis on our dataset for the first environment.
    For each scene, we show three novel views.
    }
    \label{fig:ord_nvs_examplesB}
\end{figure*}

\begin{figure*}[p]
    \centering
    \adjustimage{width=\textwidth}{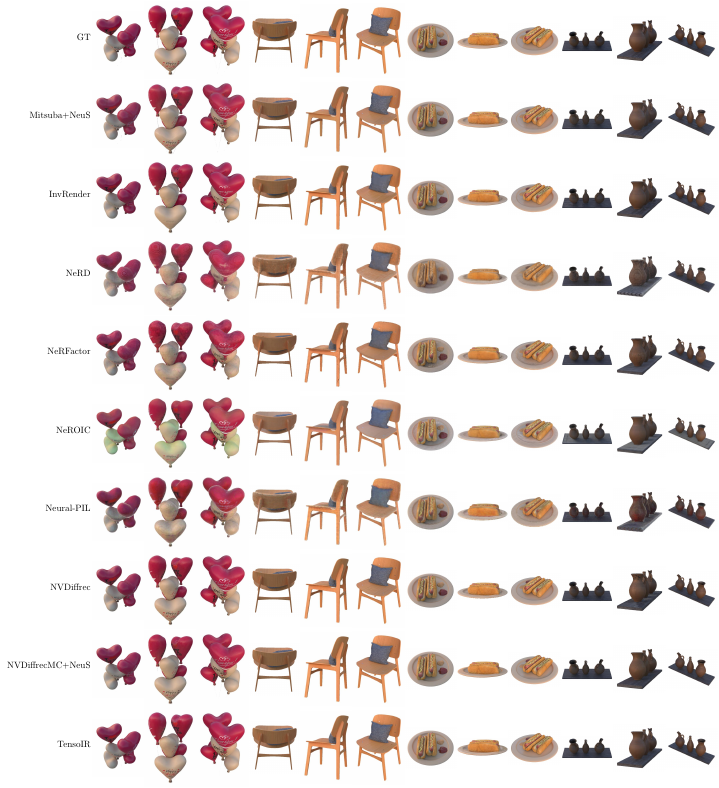}
    \caption{Qualitative examples for novel view synthesis on the Synthetic4Relight dataset for the first environment.
    For each scene, we show three novel views.
    }
    \label{fig:synth4relight_nvs_examples}
\end{figure*}

We show qualitative examples for novel view synthesis in Figures~\ref{fig:dtu_examples1},\ref{fig:dtu_examples2},\ref{fig:bmvs_examples},\ref{fig:ord_nvs_examplesA},\ref{fig:ord_nvs_examplesB},\ref{fig:synth4relight_nvs_examples}.

\begin{figure*}
   \centering
    \begin{subfigure}[t]{0.24\textwidth}
        \centering
        \adjustimage{width=\textwidth}{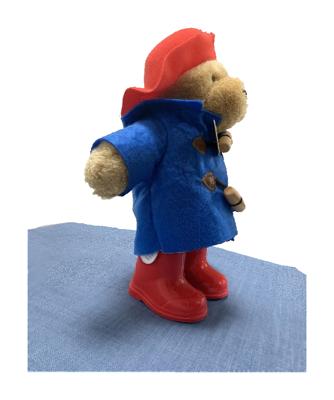}
        \caption{GT}
    \end{subfigure}
    \hfill
    \begin{subfigure}[t]{0.24\textwidth}
        \centering
        \adjustimage{width=\textwidth}{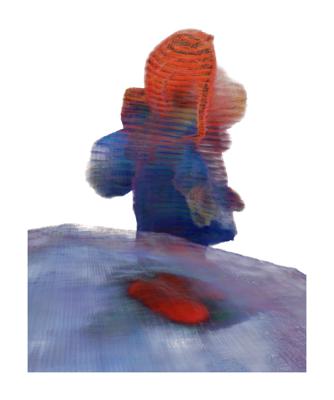}
        \caption{Neural-PIL}
    \end{subfigure}
    \hfill
    \begin{subfigure}[t]{0.20\textwidth}
        \centering
        \adjustimage{width=\textwidth}{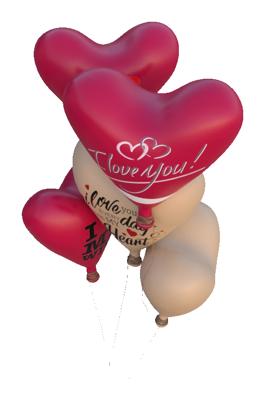}
        \caption{GT}
    \end{subfigure}
    \hfill
    \begin{subfigure}[t]{0.20\textwidth}
        \centering
        \adjustimage{width=\textwidth}{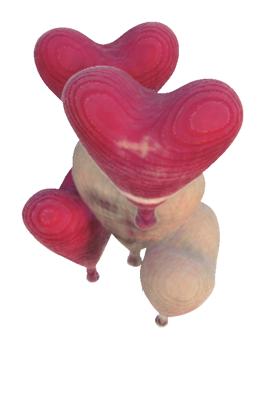}
        \caption{NeRD}
    \end{subfigure}
    \caption{Obvious failure cases that can be observed irrespective of the task. \textbf{(a),(b)} show an obvious problem in the shape. \textbf{(c),(d)} show the lack of texture detail.}
    \label{fig:nvs-problems}
\end{figure*}

We can observe obvious failures with the shape and shading for this task too but miss deficiencies like missing or baked shadows, or problems with the material parameters.
We show examples of failures in \ref{fig:nvs-problems}.

%% file: appendix_datasheet.tex
\section{Datasheet}
\label{sec:datasheet}
Our datasheet follows the template in \cite{gebru2021datasheets}.

\subsection*{Motivation}

\Q{For what purpose was the dataset created?}
The dataset was created for evaluating the reconstruction of shape and material of objects for relighting in novel environments.

\Q{Who created the dataset and on behalf of which entity?}
The authors of this paper.

\Q{Who funded the creation of the dataset?}
The dataset was created as part of a research project at Intel Labs.

\subsubsection*{Composition}
\Q{What do the instances that comprise the dataset represent?}
The instances represent photos of objects from multiple view points and equirectangular projections of environments.

\Q{How many instances are there in total?}
Currently there are 8 objects each recorded in 3 environments.

\Q{Does the dataset contain all possible instances or is it a sample of instances from a larger set?}
N/A. The dataset is not sampled from a larger existing set.

\Q{What data does each instance consist of?}
Each instance contains images of an object for the reconstruction and a separate set of images and environment maps for quantitative evaluation.
For all images for reconstruction there are approximate foreground masks.
For the test images there are expert annotated foreground masks.
All images come with a pose and intrinsic camera parameters.
A coarse bounding box and exposure values for each image are given as text files.

\Q{Is there a label or target associated with each instance?}
The test images for each object are the target.

\Q{Is any information missing from individual instances?}
No.

\Q{Are relationships between individual instances made explicit?}
No.

\Q{Are there recommended data splits (e.g., training, development/validation,
testing)?}
No. The dataset is meant for testing only.

\Q{Are there any errors, sources of noise, or redundancies in the
dataset?}
The imaging and calibration process is subject to noise. No errors are known at this time. Errors will be fixed in future revisions of the dataset.

\Q{Is the dataset self-contained, or does it link to or otherwise rely on
external resources?}
The dataset is self-contained.

\Q{Does the dataset contain data that might be considered confidential?}
No.

\Q{Does the dataset contain data that, if viewed directly, might be offensive, insulting, threatening, or might otherwise cause anxiety?}
No.

\Q{Does the dataset relate to People?}
No.

\subsubsection*{Collection process}

\Q{How was the data associated with each instance acquired?}
We use a DSLR camera for the input and test images and a 360 camera for the environment maps.

\Q{What mechanisms or procedures were used to collect the data?}
We capture the data in a specific order and with specific camera settings to enable the creation of HDR images for environment maps and test images. We refer to \sectn{sec:data-collection} for more details.

\Q{If the dataset is a sample from a larger set, what was the sampling strategy?}
N/A

\Q{Who was involved in the data collection process and how were they compensated?}
Only the authors of this paper were involved in the process.

\Q{Over what timeframe was the data collected?}
The images were collected September 2022 to March 2023.

\Q{Were any ethical review processes conducted?}
No.

\Q{Does the dataset relate to People?}
No.

\subsubsection*{Preprocesssing / cleaning / labelling}

\Q{Was any preprocessing/cleaning/labeling of the data done?}
Yes, the data is preprocessed to allow researchers to focus on the task.
All calibrations are a result of this preprocessing. We refer to \sectn{sec:data-collection} for more details.

\Q{Was the “raw” data saved in addition to the preprocessed/cleaned/labeled
data?}
Yes.

\Q{Is the software that was used to preprocess/clean/label the data
available?}
Yes. Please see the source code repository.

\subsubsection*{Uses}

\Q{Has the dataset been used for any tasks already?}
Yes. We evaluated existing methods on our dataset in this paper.

\Q{Is there a repository that links to any or all papers or systems that
use the dataset?}
No. This paper is the first to use this dataset.

\Q{What (other) tasks could the dataset be used for?}
A subset of the dataset can be used to evaluate novel view synthesis for objects.

\Q{Is there anything about the composition of the dataset or the way
it was collected and preprocessed/cleaned/labeled that might impact future uses?}
No.

\Q{Are there tasks for which the dataset should not be used?}
No.

\subsubsection*{Distribution}

\Q{Will the dataset be distributed to third parties outside of the entity on behalf of which
the dataset was created?}
Yes. The dataset will be made available to the public.

\Q{How will the dataset be distributed?}
The dataset will be hosted on GitHub along with the source code used for creating the dataset.

\Q{When will the dataset be distributed?}
The dataset will be made available on publication of the accompanying paper.

\Q{Will the dataset be distributed under a copyright or other intellectual property (IP)
license, and/or under applicable terms of use (ToU)?}
The dataset will be licensed under the CDLA-permissive 2.0 license.

\Q{Have any third parties imposed IP-based or other restrictions on
the data associated with the instances?}
No.

\Q{Do any export controls or other regulatory restrictions apply to
the dataset or to individual instances?}
No.

\subsubsection*{Maintenance}

\Q{Who will be supporting/hosting/maintaining the dataset?}
The authors listed on this paper will support and maintain the dataset.

\Q{How can the owner/curator/manager of the dataset be contacted?}
The authors listed on this paper can be contacted via email or the datasets GitHub repository.

\Q{Is there an erratum?}
No.

\Q{Will the dataset be updated?}
Yes. The dataset will be updated with new objects to adapt its difficulty to push the state-of-the-art.

\Q{Will older versions of the dataset continue to be supported/hosted/maintained?}
Yes. The data will be versioned.

\Q{If others want to extend/augment/build on/contribute to the dataset, is there a mechanism for them to do so?}
Yes. The code used for building the dataset in available in the datasets GitHub repo and others can contact the authors for adding new objects.